\documentclass[useAMS,usenatbib,referee]{biom}
\usepackage{amsmath}
\usepackage{mathtools}
\usepackage{xfrac}
\usepackage{psfrag,epsf}
\usepackage{url} 
\usepackage{float}
\usepackage[T1]{fontenc}
\usepackage{colortbl}
\usepackage{rotating}
\usepackage{graphicx}

\usepackage{amssymb}

\usepackage[compact]{titlesec}
\titlespacing{\section}{0pt}{2ex}{1ex}
\titlespacing{\subsection}{0pt}{1ex}{0ex}
\titlespacing{\subsubsection}{0pt}{0.5ex}{0ex}

\DeclareMathOperator{\E}{\mathbb{E}}
\DeclarePairedDelimiter\abs{\lvert}{\rvert}%
\DeclarePairedDelimiter\norm{\lVert}{\rVert}%

\def\bSig\mathbf{\Sigma}

\makeatletter
\let\oldabs\abs
\def\abs{\@ifstar{\oldabs}{\oldabs*}}
\let\oldnorm\norm
\def\norm{\@ifstar{\oldnorm}{\oldnorm*}}
\makeatother

\title[NIMIWAE]{Unsupervised Imputation of Non-ignorably Missing Data Using Importance-Weighted Autoencoders}

\author{David K. Lim$^{1,*}$\email{deelim@live.unc.edu},
Naim U. Rashid$^{1,**}$\email{naim@unc.edu}, and
Junier B. Oliva$^{2,***}$\email{joliva@cs.unc.edu}, and
Joseph G. Ibrahim$^{1,****}$\email{ibrahim@bios.unc.edu} \\
$^{1}$Department of Biostatistics, University of North Carolina at Chapel Hill, Chapel Hill, NC, USA\\
$^{2}$Department of Computer Science, University of North Carolina at Chapel Hill, Chapel Hill, NC, USA}

\begin{document}

\pagerange{\pageref{firstpage}--\pageref{lastpage}}

\label{firstpage}

\begin{abstract}   

Deep Learning (DL) methods have dramatically increased in popularity in recent years. While its initial success was demonstrated in the classification and manipulation of image data, there has been significant growth in the application of DL methods to problems in the biomedical sciences. However, the greater prevalence and complexity of missing data in biomedical datasets present significant challenges for DL methods. Here, we provide a formal treatment of missing data in the context of Variational Autoencoders (VAEs), a popular unsupervised DL architecture commonly utilized for dimension reduction, imputation, and learning latent representations of complex data. We propose a new VAE architecture, NIMIWAE, that is one of the first to flexibly account for both ignorable and non-ignorable patterns of missingness in input features at training time. Following training, samples can be drawn from the approximate posterior distribution of the missing data can be used for multiple imputation, facilitating downstream analyses on high dimensional incomplete datasets. We demonstrate through statistical simulation that our method outperforms existing approaches for unsupervised learning tasks and imputation accuracy. We conclude with a case study of an EHR dataset pertaining to 12,000 ICU patients containing a large number of diagnostic measurements and clinical outcomes, where many features are only partially observed.


\end{abstract}


\maketitle



\section{Introduction} \label{sec:intro}


Deep Learning (DL) methods have dramatically increased in popularity in recent years.  While its initial success was demonstrated in the classification and manipulation of image data, there has been significant growth in the application of DL methods to an array of problems in the biomedical sciences \citep{Shickel2018,Lopez2018}. The presence of complex interactions among high-dimensional features in modern biomedical data has motivated the use of Variational Autoencoders (VAEs), a DL architecture commonly utilized for unsupervised learning tasks such as dimension reduction, representational learning, and generation of synthetic data mimicking real input data, which may be unavailable due to patient confidentiality 
\citep{Shickel2018}. In prior evaluations, VAEs have shown tremendous performance in data generation and representation learning \citep{Kingma2019}.

Some of the useful characteristics of the VAE can be derived from its underlying architecture. This structure is similar to that of the Autoencoder (AE), which consists of two major parts. The first part is the encoder neural network, which in the AE deterministically transforms the input data into a lower-dimensional set of factors that is assumed to capture the salient qualities of the data. The second is the decoder, which is a separate neural network that attempts to reconstruct the original data from the lower dimensional space itself \citep{Tschannen2018}. The quality of the dimension reduction and reconstruction is measured by the reconstruction error, which is measured by the difference in the input data and its reconstructed version.  This lower dimensional space may be used to define structure in the underlying set of input data \citep{Wang2016}, similar to PCA. However, the encoder and decoder are often represented by feed forward neural networks, allowing for complex non-linear interactions between features to be captured in both \citep{LeCun2015}. The VAE additionally imposes a probabilistic assumption on the input data and the lower-dimensional (or "latent") space \citep{Doersch2016}, where explicit encoder and decoder distributions are defined, and neural networks are used to output the parameters of each distribution \citep{Kingma2019}. The learned decoder allows the VAE to also generate synthetic data that has high fidelity to the original training data, given the learned latent space. %

However, accounting for the prevalence and complexity of missing data in modern biomedical datasets presents a significant challenge for training VAEs and other DL architectures. Existing DL methods that claim to handle missing data either require pre-training on studies with fully observed data, utilize heuristics to replace missing observations, or cannot handle more complicated forms of missingness, such as Missing at Random (MAR), or Missing Not a Random (MNAR) \citep{Li2020, Strauss2021}.  Intuitively, when observations with missing values differ systematically from those without missingness, results from standard approaches that do not properly account for this fact may no longer be accurate. For example, the common presence of missing data across patient records in electronic health record data has been shown to present a significant barrier to the generalizability and applicability of deep learning methods \citep{Wells2013}. EHR data also typically contain many features pertaining to a large number of patients \citep{Ross2014}, and such features may show complex interactions with each other as well as with clinical outcomes of interest \citep{Vinzamuri2013}, compounding the impact of the missingness of these features. For these reasons, the ability to flexibly account for different patterns of missing data in modern biomedical datasets, especially in the more difficult non-ignorable missingness setting, would have great utility.  
In this manuscript, we introduce a novel deep learning architecture called Non-Ignorably Missing Importance-weighted Autoencoder (NIMIWAE) that treats missing observations as latent variables in the VAE framework using Importance-Weighted Autoencoders (IWAEs). Here we model the probability of missingness using a feed forward neural network, facilitating more flexible handling of MNAR patterns of missingness across a large set of available features. Through simulations, we show that proper modelling of the missingness mechanism increases the accuracy of missing data imputation, as well as downstream coefficient estimation using these imputed datasets. Lastly, we show that in the Physionet 2012 Challenge EHR dataset, accounting for MNAR missingness results in significant differences in downstream fitted models to predict in-hospital mortality. More generally, NIMIWAE is a flexible framework within the VAE family to handle multiple patterns of missingness commonly found in complex biomedical datasets.



\section{Methods} \label{sec:meth}

\subsection{Variational Autoencoder} \label{sec:VAE}
Let $\mathbf{X}$ be an $n \times p$ data matrix, where $\mathbf{x}_i$ denotes the observation vector pertaining to the $i^{th}$ observation, $i=1,\ldots,n$, and $x_{ij}$ denotes the value of the $j^{th}$ feature in this vector, $j=1,\ldots,p$. In a VAE, we assume $\mathbf{x}_1,\ldots,\mathbf{x}_n$ are i.i.d. samples from a multivariate p.d.f or ``generative model" $p_\psi(\mathbf{X} | \mathbf{Z})$.  Here, $\mathbf{Z}$ is an $n \times d$ matrix, such that $\mathbf{Z}=\{\mathbf{z}_1,\cdots,\mathbf{z}_n\}$ and $\mathbf{z}_i$ is a latent vector of length $d$ pertaining to the $i^{th}$ sample  \citep{Kingma2019}. Typically it is assumed that $d \leq p$, such that $\mathbf{Z}$ constitutes a lower-dimensional representation of the original data $\mathbf{X}$.  The parameters $\psi$ and conditional distribution  $p_\psi(\mathbf{X} | \mathbf{Z})$ indicate how the observed data $\mathbf{X}$ may be generated from $\mathbf{Z}$. In this manner, a VAE aims to learn accurate representations of high-dimensional data, and may be used to generate synthetic data with similar qualities as its training data. These aspects are also aided through the use of embedded deep learning neural networks, for example within $p_\psi(\mathbf{X} | \mathbf{Z})$, which also facilitate its applicability to larger dimensions and complex datasets.

\subsubsection{Objective Function} \label{sec:VAEobj}
Since $\psi$ is unknown, learning is performed by maximizing the marginal log-likelihood of $\mathbf{X}$ with respect to $\psi$,  where we denote this marginal log-likelihood as $\log p_\psi(\mathbf{X})=\log \int p_\psi(\mathbf{X},\mathbf{Z}) d\mathbf{Z} = \log \int p_\psi(\mathbf{X} | \mathbf{Z})p(\mathbf{Z}) d\mathbf{Z}$. However, due to the integral involved, this quantity is often intractable and is difficult to maximize directly. Therefore, VAE's alternatively optimize the so-called ``Evidence Lower Bound" (ELBO), which has the following form \citep{Kingma2019}:
\begin{align*}
\mathcal{L}^{ELBO}(\theta,\psi) &= \E_{\mathbf{Z} \sim q_\theta(\mathbf{Z}|\mathbf{X})} \log \left[ \frac{p_\psi(\mathbf{X}|\mathbf{Z})p(\mathbf{Z})}{q_\theta(\mathbf{Z}|\mathbf{X})} \right]\stepcounter{equation}\tag{\theequation}\label{eqn:ELBO1} \\
\hat{\mathcal{L}}^{ELBO}_K(\theta,\psi) &= \frac{1}{K} \sum_{k=1}^K \log \left[ \frac{p_\psi(\mathbf{X}|\tilde{\mathbf{Z}}_k)p(\tilde{\mathbf{Z}}_k)}{q_\theta(\tilde{\mathbf{Z}}_k|\mathbf{X})} \right].\stepcounter{equation}\tag{\theequation}\label{eqn:ELBO2}
\end{align*}
Here, $\mathcal{L}^{ELBO}(\theta,\psi)$ denotes the ELBO such that $\mathcal{L}^{ELBO}(\theta,\psi) \leq \log p_\psi(\mathbf{X})$.   Also let $\hat{\mathcal{L}}^{ELBO}_K(\theta,\psi)$ denote the empirical approximation to Eq. (\ref{eqn:ELBO1}) computed by Monte Carlo integration, such that  $\mathcal{L}^{ELBO}(\theta,\psi) \approx \hat{\mathcal{L}}_K^{ELBO}(\theta,\psi)$ and  $\tilde{\mathbf{Z}}_1,\ldots,\tilde{\mathbf{Z}}_K$ are $K$ samples drawn from $q_\theta(\mathbf{Z}|\mathbf{X})$, the variational approximation of the true but intractable posterior $p_\psi(\mathbf{Z}|\mathbf{X})$, also called the ``recognition model". Furthermore,  denote $f_\psi(\mathbf{Z})$ and $g_\theta(\mathbf{X})$ as the decoder and encoder feed forward neural networks of the VAE, where $\psi$ and $
\theta$ are the sets of weights and biases pertaining to each of these neural networks, respectively. Given $\mathbf{Z}$, $f_\psi(\mathbf{Z})$ outputs the distributional parameters pertaining to $p_\psi(\mathbf{X}|\mathbf{Z})$.  Given $\mathbf{X}$, $g_\theta(\mathbf{X})$ outputs the distributional parameters for $q_\theta(\mathbf{Z}|\mathbf{X})$. 

In variational inference, $q_\theta(\mathbf{Z}|\mathbf{X})$ is constrained to be from a class of simple distributions, or ``variational family", to obtain the best candidate from within that class to approximate $p_\psi(\mathbf{Z}|\mathbf{X})$. Variational inference is usually used in tandem with amortization  of the parameters where the neural network parameters are shared across observations \citep{Gershman2014}, allowing for stochastic gradient descent (SGD) to be used for optimization of Eq. (\ref{eqn:ELBO2}) \citep{Kingma2019}. In practice, both $q_\theta(\mathbf{Z}|\mathbf{X})$ and $p(\mathbf{Z})$ are typically assumed to have simple forms, such as multivariate Gaussians with diagonal covariance structures, and $q_\theta(\mathbf{Z}|\mathbf{X})$ is commonly assumed to be factorizable, such that $q_\theta(\mathbf{Z}|\mathbf{X}) = \prod_{i=1}^n q_\theta(\mathbf{z}_i|\mathbf{x}_i)$ \citep{Kingma2019}.

\subsubsection{Estimation Procedure and Use Cases}
Let $(\hat{\theta}^{(t)},\hat{\psi}^{(t)})$ be the estimates of $(\theta,\psi)$ at update (or iteration) $t$. For $t=0$, these values are often initialized to small values centered around 0, although other initialization schemes may be used \citep{Saxe2014,Murphy2016}.  Each subsequent update $t\geq 1$ consists of two general steps to maximize $\mathcal{L}(\theta,\psi)$. First, $K$ samples are drawn from $q_{\hat{\theta}^{(t)}}(\mathbf{Z}|\mathbf{X})$ to compute the quantity in Eq. (\ref{eqn:ELBO2}), conditional on $(\hat{\theta}^{(t)},\hat{\psi}^{(t)})$. Then, the so-called ``reparametrization trick" is utilized to facilitate the calculation of gradients of this approximation to obtain $(\hat{\theta}^{t+1},\hat{\psi}^{t+1})$ using stochastic gradient descent \citep{Kingma2019}. This procedure may be repeated for a fixed number of iterations, or may be terminated early via pre-specified early stop criteria \citep{Prechelt1998}. \citet{Kingma2019} provides additional details on the maximization procedure for VAEs.  The networks $f_\psi(\mathbf{Z})$ and $g_\theta(\mathbf{X})$ also allow the VAE to capture complex and non-linear relationships between features when outputting the distributional parameters for the generative and recognition models, respectively, through the inclusion of hidden layers in each network. The number of hidden layers and nodes per layer for each network are commonly determined via hyperparameter tuning. 

After model fitting, the VAE has several useful features.  First, synthetic data can be generated by sampling from the learned generative model $p_{\hat{\psi}}(\mathbf{X}|\mathbf{Z})$ after drawing a sample from $q_{\hat{\theta}}(\mathbf{Z}|\mathbf{X})$ \citep{Mattei2019,Nazabal2018}. Second, the posterior modes in the latent space $\mathbf{Z}$ can be determined from  $q_{\hat{\theta}}(\mathbf{Z}|\mathbf{X})$.  These posterior modes may be used for purposes such as clustering or substructure discovery \citep{Lim2020}.  In some applications, $p_{\hat{\psi}}(\mathbf{X}|\mathbf{Z})$ may also be used to directly perform imputation \citep{Nazabal2018}, however the statistical properties of this procedure have not been thoroughly discussed in prior work.  

 \subsection{Importance-Weighted Autoencoder} \label{sec:IWAE}
The IWAE \citep{Burda2015} is a generalization of the standard VAE, where the resulting IWAE bound, corresponding to the ELBO in Eq. (\ref{eqn:ELBO1}), can be written as
\begin{align*}
\mathcal{L}_K^{IWAE}(\theta,\psi)&= \E_{\mathbf{Z}_k \sim q_\theta(\mathbf{Z}|\mathbf{X})} \log \left[ \frac{1}{K} \sum_{k=1}^K \frac{p_\psi(\mathbf{X}|\mathbf{Z}_k)p(\mathbf{Z}_k)}{q_\theta(\mathbf{Z}_k|\mathbf{X})} \right] \stepcounter{equation}\tag{\theequation}\label{eqn:IWAEbound1}\\
\hat{\mathcal{L}}_K^{IWAE}(\theta,\psi)&= \log \left[ \frac{1}{K} \sum_{k=1}^K \frac{p_\psi(\mathbf{X}|\tilde{\mathbf{Z}}_k)p(\tilde{\mathbf{Z}}_k)}{q_\theta(\tilde{\mathbf{Z}}_k|\mathbf{X})} \right].\stepcounter{equation}\tag{\theequation}\label{eqn:IWAEbound2}
\end{align*}
An important distinction in Eq. (\ref{eqn:IWAEbound1}) from Eq. (\ref{eqn:ELBO1}) is that a VAE assumes a single latent variable $\mathbf{Z}$ in Eq. (\ref{eqn:ELBO1}) that is sampled $K$ times in the ELBO approximation from Eq. (\ref{eqn:ELBO2}).  In contrast, an IWAE assumes $K$ i.i.d. latent variables in the expression for its lower bound, where $\mathbf{Z}_1,\ldots,\mathbf{Z}_K \stackrel{i.i.d}{\sim} q_{\theta}(\mathbf{Z}|\mathbf{X})$. The contribution of each $\mathbf{Z}_k$ in Eq. (\ref{eqn:IWAEbound1}) is weighted by $\frac{p(\mathbf{Z}_k)}{q_\theta(\mathbf{Z}_k|\mathbf{X})}$. Then, to compute its empirical approximation  $\hat{\mathcal{L}}_K^{IWAE}$, typically only one sample is drawn for each $\mathbf{Z}_k$ in Eq. (\ref{eqn:IWAEbound2}).  For $K>1$, \citet{Burda2015} showed that $\log p(\mathbf{X}) \geq \mathcal{L}_{K+1}^{IWAE} \geq \mathcal{L}_K^{IWAE}$, such that $\mathcal{L}_K^{IWAE} \rightarrow \log p(\mathbf{X})$ as $K \rightarrow \infty$ if $p_\psi(\mathbf{X},\mathbf{Z})/q_\theta(\mathbf{Z}|\mathbf{X})$ is bounded. Thus, the IWAE bound more closely approximates the true marginal log likelihood when $K>1$ \citep{Cremer2017}, at the cost of greater computational burden. If $K=1$, $\mathcal{L}_1^{IWAE}=\mathcal{L}^{VAE}$, and the IWAE corresponds exactly to the standard VAE. In this way, the IWAE can be considered to be part of the VAE family, and we refer to methods that use either VAEs or IWAEs broadly as ``VAE methods".  A visualization of the workflow for an IWAE (without missing data) can be found in Web Appendix A. 

VAE methods have shown excellent performance in representation learning on many types of data. However, the presence of missingness in $\mathbf{X}$ presents significant challenges to the above modeling procedures and the application of VAE methods in general.

\subsection{Missing Data} \label{sec:miss}
In this section, we first introduce notation for missing data and review the different mechanisms of missingness, as described in the statistical literature. Let the data be factored such that $\mathbf{X}=\{\mathbf{X}^o,\mathbf{X}^m\}$, with $\mathbf{X}^o$ denoting the observed values and $\mathbf{X}^m$ denoting the missing values. For each observation vector $\mathbf{x}_i$, denote $\mathbf{x}_i^o$ and $\mathbf{x}_i^m$ respectively to be the observed and missing features of $\mathbf{x}_i$. Also, let $\mathbf{R}$ be a matrix of the same dimension as $\mathbf{X}$, with entries $r_{ij}=I(x_{ij}$ is observed$)$ for the $i^{th}$ observation and $j^{th}$ feature, where $I(\cdot)$ denotes the indicator function. In this way, $\mathbf{R}$ is the ``mask" matrix pertaining to $\mathbf{X}$, such that $\mathbf{x}_i^o=\{x_{ij}:r_{ij}=1\}$ and $\mathbf{x}_i^m=\{x_{ij}:r_{ij}=0\}$ for all $i=1,\ldots,n$ and $j=1,\ldots,p$.

Missingness was classified into three major categories, or mechanisms, in the seminal work by \citet{Little2002}. These mechanisms are missing completely at random (MCAR), missing at random (MAR), and missing not at random (MNAR), and they satisfy the following relations: (1) MCAR: $p(\mathbf{r}_{i}|\mathbf{x}_i,\mathbf{z}_i,\boldsymbol{\phi}) = p(\mathbf{r}_i|\boldsymbol{\phi})$, (2) MAR: $p(\mathbf{r}_{i}|\mathbf{x}_i,\mathbf{z}_i,\boldsymbol{\phi}) = p(\mathbf{r}_i|\mathbf{x}_i^o,\boldsymbol{\phi})$, and (3) MNAR: $p(\mathbf{r}_{i}|\mathbf{x}_i,\mathbf{z}_i,\boldsymbol{\phi}) = p(\mathbf{r}_i|\mathbf{x}_i^o,\mathbf{x}_i^m,\mathbf{z}_i,\boldsymbol{\phi})$. Here, $\boldsymbol{\phi}$ denotes the unknown parameters pertaining to the missingess model $p(\mathbf{r}_{i}|\mathbf{x}_i,\mathbf{z}_i,\boldsymbol{\phi})$, where $\mathbf{r}_i=\{r_{i1},\ldots,r_{ip}\}$. We discuss forms of this model in Section \ref{sec:NI_miss}. In the presence of missingness mask $\mathbf{R}$, the marginal log-likelihood may be written as
\begin{equation}
\log p_{\psi,\phi}(\mathbf{X}^o,\mathbf{R})=\log \iint p_{\psi,\phi}(\mathbf{X}^o,\mathbf{X}^m,\mathbf{Z},\mathbf{R}) d\mathbf{X}^m d\mathbf{Z}.
\label{eqn:LLNonignorable}
\end{equation}

We may factor $p_{\psi,\phi}(\mathbf{X}^o,\mathbf{X}^m,\mathbf{Z},\mathbf{R})$ using the selection model factorization \citep{Diggle1994}, which is written as $p_{\psi,\phi}(\mathbf{X}^o,\mathbf{X}^m,\mathbf{Z},\mathbf{R})=p_\psi(\mathbf{X}^o,\mathbf{X}^m|\mathbf{Z})p(\mathbf{Z})p(\mathbf{R}|\mathbf{X},\mathbf{Z},\boldsymbol{\phi})$, where $p(\mathbf{R}|\mathbf{X},\mathbf{Z},\boldsymbol{\phi})=\prod_{i=1}^{n} p(\mathbf{r}_i|\mathbf{x}_i,\mathbf{z}_i,\boldsymbol{\phi})$. 



\subsubsection{Ignorable Missingness} \label{sec:I_miss}

In a likelihood-based analysis under either MCAR or MAR, the missingness mechanism is considered to be ``ignorable" such that the missingness mechanism need not be explicitly modelled in these cases \citep{Rubin1976,Little2002}. Under ignorable missingness, the left hand side of Eq. (\ref{eqn:LLNonignorable}) can be separated into $\log p_\psi(\mathbf{X}^o)+ \log p_\phi(\mathbf{R}|\mathbf{X}^o)$, where $p_\psi(\mathbf{X}^o)$ is the marginal distribution of $\mathbf{X}^o$. Therefore, $p_\phi(\mathbf{R}|\mathbf{X}^o)$ need not be specified because inference on the parameters of interest pertaining to $p_\psi(\mathbf{X}^o)$ is independent of $p_\phi(\mathbf{R}|\mathbf{X}^o)$. Then, one aims to maximize the quantity
\begin{equation}
\log p_\psi(\mathbf{X}^o)=\log \iint p_\psi(\mathbf{X}^o,\mathbf{X}^m,\mathbf{Z}) d\mathbf{X}^m d\mathbf{Z} = \log \int p_\psi(\mathbf{X}^o,\mathbf{Z}) d\mathbf{Z}.
\label{eqn:LLIgnorable}
\end{equation}
This quantity can be bounded below exactly as in Section \ref{sec:VAEobj}, conditioning on just the observed data $\mathbf{X}^o$, rather than the full data $\mathbf{X}$. Existing methods typically take advantage of this simplification, and are shown to perform well under ignorable missingness. Details for these methods are given in Web Appendix A.

\subsubsection{Non-ignorable Missingness} \label{sec:NI_miss}
In contrast, non-ignorable (or MNAR) missingness refers to the case where the missingness can be dependent on any unobserved values, including the missing entries $\mathbf{x}_i^m$. MNAR missingness can also be dependent on $\mathbf{x}_i^o$ as well as latent values like $\mathbf{Z}$, and thus MNAR represents the most general and difficult case of missingness in practice. Here, we assume that $\mathbf{R}$ is independent of $\mathbf{Z}$, as conditioning on such latent factors may be computationally redundant based on the assumed data generating process \citep{Ibrahim2001}. In this setting, the missingness typically requires specification of a model for the missingness $p(\mathbf{R}|\mathbf{X},\boldsymbol{\phi})$ \citep{Stubbendick2003}. Current VAE methods are only able to handle MCAR or MAR missingness, and there is no method to properly deal with the more difficult MNAR case. This issue is especially problematic because missingness in many real world applications have been posited to be non-ignorable \citep{Beaulieu-Jones2016,OShea2019}.

There have been a number of ways to specify $p(\mathbf{R}|\mathbf{X},\boldsymbol{\phi})$ in statistical literature. For example, \citet{Diggle1994} proposes a binomial model for the missing data mechanism: $$p(\mathbf{R}|\mathbf{X},\boldsymbol{\phi})=\prod_{i=1}^{n} \prod_{j_m=1}^{p} \left[ p(r_{ij_m}=1|\mathbf{x}_i,\boldsymbol{\phi}_{j_m})\right]^{r_{ij_m}}\left[ 1-p(r_{ij_m}=1|\mathbf{x}_i,\boldsymbol{\phi}_{j_m})\right]^{1-r_{ij_m}},$$ where $j_m=1,\ldots,p_{miss}$ indexes the $p_{miss}$ features in $\mathbf{X}$ that contain missingness, $\boldsymbol{\phi}_{j_m}$ is the sets of coefficients pertaining to $j_m^{th}$ missingness model, and $p(r_{ij_m}=1|\mathbf{x}_i,\boldsymbol{\phi}_{j_m})$ can be modeled straightforwardly by a logistic regression model, such that \begin{equation}
logit[p(r_{ij_m}=1|\mathbf{x}_i,\boldsymbol{\phi}_{j_m})]=\phi_{0j_m}+\mathbf{x}_i^o\boldsymbol{\phi}_{1j_m}+\mathbf{x}_i^m\boldsymbol{\phi}_{2j_m},\label{eqn:logistic}
\end{equation} 
where $\phi_{0j_m}$ is the intercept of the $j_m^{th}$ missingness model,  $\boldsymbol{\phi}_{1j_m}=\{\phi_{1,j_m,1},\ldots,\phi_{1,j_m,p_{obs}}\}^T$ is a  $p_{obs} \times 1$ vector of coefficients pertaining to the fully-observed features, and $\boldsymbol{\phi}_{2j_m}=\{\phi_{2,j_m,1},\ldots,\phi_{2,j_m,p_{miss}}\}^T$ is a  $p_{miss} \times 1$ vector of coefficients pertaining to the missing features. 

\subsection{NIMIWAE: IWAE with Nonignorable Missingness} \label{sec:VAENonignorable}

We now propose a novel method to perform statistical learning and imputation using an IWAE in the presence of missing data (NIMIWAE), assuming missingness is nonignorable. We later show how this model can be simplified when missingness is assumed to be ignorable (IMIWAE).  First, we specify a general form of the lower bound, in which we utilize the general IWAE framework to form a tighter bound on the marginal log-likelihood than the VAE ELBO.  Let us define $q_\theta(\mathbf{Z},\mathbf{X}^m)$ as the variational joint posterior pertaining to $(\mathbf{Z},\mathbf{X}^m)$. We can factor this variational joint posterior as $q_\theta(\mathbf{Z},\mathbf{X}^m) = q_{\theta_1}(\mathbf{Z}|\mathbf{X}^o)q_{\theta_2}(\mathbf{X}^m|\mathbf{Z},\mathbf{X}^o,\mathbf{R})$. Here, for $k=1,\ldots,K$, we assume $\mathbf{Z}_k \stackrel{i.i.d}{\sim} q_{\theta_1}(\mathbf{Z}|\mathbf{X}^o)$ similar to the traditional IWAE. We additionally assume i.i.d. latent variables $\mathbf{X}_{k}^m \stackrel{i.i.d}{\sim} q_{\theta_2}(\mathbf{X}^m|\mathbf{Z},\mathbf{X}^o,\mathbf{R})$ pertaining to the missing features, where each $\mathbf{X}_{k}^m$ has dimensionality $p_{miss}$. Similar to traditional VAEs, we utilize the class of factorized variational posteriors, except now $q_{\theta}(\mathbf{Z},\mathbf{X}^m)=\prod_{i=1}^{n}q_{\theta}(\mathbf{z}_i,\mathbf{x}_i^m)$ and  $q_{\theta}(\mathbf{z}_i,\mathbf{x}_i^m)=q_{\theta_1}(\mathbf{z}_i|\mathbf{x}_i^o)q_{\theta_2}(\mathbf{x}_i^m|\mathbf{z}_i,\mathbf{x}_i^o,\mathbf{r}_i)$. Then, denoting $\mathbf{z}_{ik}$ and $\mathbf{x}_{ik}^m$ as the $i^{th}$ observation vectors of $\mathbf{Z}_k$ and $\mathbf{X}_k^m$, respectively, we have $\mathbf{z}_{i1},\ldots,\mathbf{z}_{iK} \stackrel{i.i.d}{\sim} q_{\theta_1}(\mathbf{z}_i|\mathbf{x}_i^o)$ and $\mathbf{x}_{i1}^m,\ldots,\mathbf{x}_{iK}^m \stackrel{i.i.d}{\sim} q_{\theta_2}(\mathbf{x}_i^m|\mathbf{z}_i,\mathbf{x}_i^o,\mathbf{r}_i)$. The form of the lower bound, which we call the \textbf{N}on\textbf{I}gnorably \textbf{M}issing \textbf{I}mportance-\textbf{W}eighted \textbf{A}uto \textbf{E}ncoder bound, or ``\textbf{NIMIWAE} bound'', is derived as follows:
\begin{align*}
\log p_{\psi,\phi}(\mathbf{X}^o,\mathbf{R}) &= \sum_{i=1}^{n} \log p_{\psi,\phi}(\mathbf{x}_i^o,\mathbf{r}_i) \\
&= \sum_{i=1}^n \log \left[\iint p_{\psi,\phi}(\mathbf{x}_i^o,\mathbf{x}_i^m,\mathbf{r}_i,\mathbf{z}_i) d\mathbf{z}_i d\mathbf{x}_i^m \right]\\
&= \sum_{i=1}^n \log \E_{(\mathbf{z}_{ik},\mathbf{x}_{ik}^m) \sim q_{\theta}(\mathbf{z}_i,\mathbf{x}_i^m)} \left[ \frac{1}{K}\sum_{k=1}^K  \frac{p_{\psi,\phi}(\mathbf{x}_i^o,\mathbf{x}_{ik}^m,\mathbf{r}_i,\mathbf{z}_{ik})}{q_{\theta_1}(\mathbf{z}_{ik}|\mathbf{x}_i^o)q_{\theta_2}(\mathbf{x}_{ik}^m|\mathbf{z}_{ik},\mathbf{x}_i^o,\mathbf{r}_i)} \right]\\
&\geq \sum_{i=1}^{n} \E_{(\mathbf{z}_{ik},\mathbf{x}_{ik}^m) \sim q_{\theta}(\mathbf{z}_i,\mathbf{x}_i^m)} \log{ \left[ \frac{1}{K}\sum_{k=1}^{K}\frac{p_{\psi,\phi}(\mathbf{x}_i^o,\mathbf{x}_{ik}^m,\mathbf{r}_i,\mathbf{z}_{ik})}{q_{\theta_1}(\mathbf{z}_{ik}|\mathbf{x}_i^o)q_{\theta_2}(\mathbf{x}_{ik}^m|\mathbf{z}_{ik},\mathbf{x}_i^o,\mathbf{r}_i)} \right] } = \mathcal{L}_{K}^{NIMIWAE}. \stepcounter{equation}\tag{\theequation}\label{eqn:NIMIWAEbound1}
\end{align*}

As explained in Section \ref{sec:miss}, we use the selection model factorization of the joint distribution of $\{\mathbf{x}_i,\mathbf{r}_i,\mathbf{z}_i\}$, such that  $p_{\psi,\phi}(\mathbf{x}_i^o,\mathbf{x}_i^m,\mathbf{r}_i,\mathbf{z}_i)=p_{\psi}(\mathbf{x}_i^o,\mathbf{x}_i^m|\mathbf{z}_i)p(\mathbf{z}_i)p_\phi(\mathbf{r}_i|\mathbf{x}_i^o,\mathbf{x}_i^m).$ Here, $\psi$ denotes the weights and biases of the encoder and decoder neural networks, and $\phi$ denotes the weights and biases of the missingness network that learns the parameters of the missingness model.

Applying the above factorizations to Eq. (\ref{eqn:NIMIWAEbound1}), and estimating the expectations in Eq. (\ref{eqn:NIMIWAEbound1}) by sampling from $q_\theta(\mathbf{z}_i,\mathbf{x}_i^m)$ we obtain the estimate of the NIMIWAE bound:
\begin{equation}
\hat{\mathcal{L}}_{K}^{NIMIWAE}=\sum_{i=1}^{n} \log{\left[ \frac{1}{K} \sum_{k=1}^{K}  \frac{p_{\psi}(\mathbf{x}_i|\tilde{\mathbf{z}}_{ik})p(\tilde{\mathbf{z}}_{ik})p_{\phi}(\mathbf{r}_i|\mathbf{x}_i^o,\tilde{\mathbf{x}}_{ik}^m)}{q_{\theta_1}(\tilde{\mathbf{z}}_{ik}|\mathbf{x}_i^o) q_{\theta_2}(\tilde{\mathbf{x}}_{ik}^m|\tilde{\mathbf{z}}_{ik},\mathbf{x}_{i}^o,\mathbf{r}_i)} \right ] }, 
\label{eqn:NIMIWAEbound2}
\end{equation}
where $\{\tilde{\mathbf{z}}_{ik}, \tilde{\mathbf{x}}_{ik}^{m}\}$ are samples of $\{\mathbf{z}_i,\mathbf{x}_i^m\}$ that are drawn via ancestral sampling \citep{Bishop2006} from $q_{\theta_1}(\tilde{\mathbf{z}}_{ik}|\mathbf{x}_i^o)$ and $q_{\theta_2}(\tilde{\mathbf{x}}_{ik}^m|\tilde{\mathbf{z}}_{ik},\mathbf{x}_{i}^o,\mathbf{r}_i)$, respectively, and the NIMIWAE bound is optimized using the Adam optimizer \citep{Kingma2014}. In NIMIWAE, we have four neural networks $f_\psi(\mathbf{z}_i)$, $g_{\theta_1}(\mathbf{x}_i^o)$, $g_{\theta_2}(\mathbf{x}_i^o,\mathbf{r}_i,\mathbf{z}_i)$, and $h_{\phi}(\boldsymbol{x}_i)$, which respectively output the parameters pertaining to $p_\psi(\mathbf{x}_i|\mathbf{z}_{ik})$, $q_{\theta_1}(\mathbf{z}_{ik}|\mathbf{x}_i^o)$, $q_{\theta_2}(\mathbf{x}_{ik}^m|\mathbf{z}_{ik},\mathbf{x}_i^o,\mathbf{r}_i)$, and $p_\phi(\mathbf{r}_i|\mathbf{x}_i^o,\mathbf{x}_{ik}^m)$.

\begin{figure}[H]
\begin{center}
\includegraphics[width=150mm]{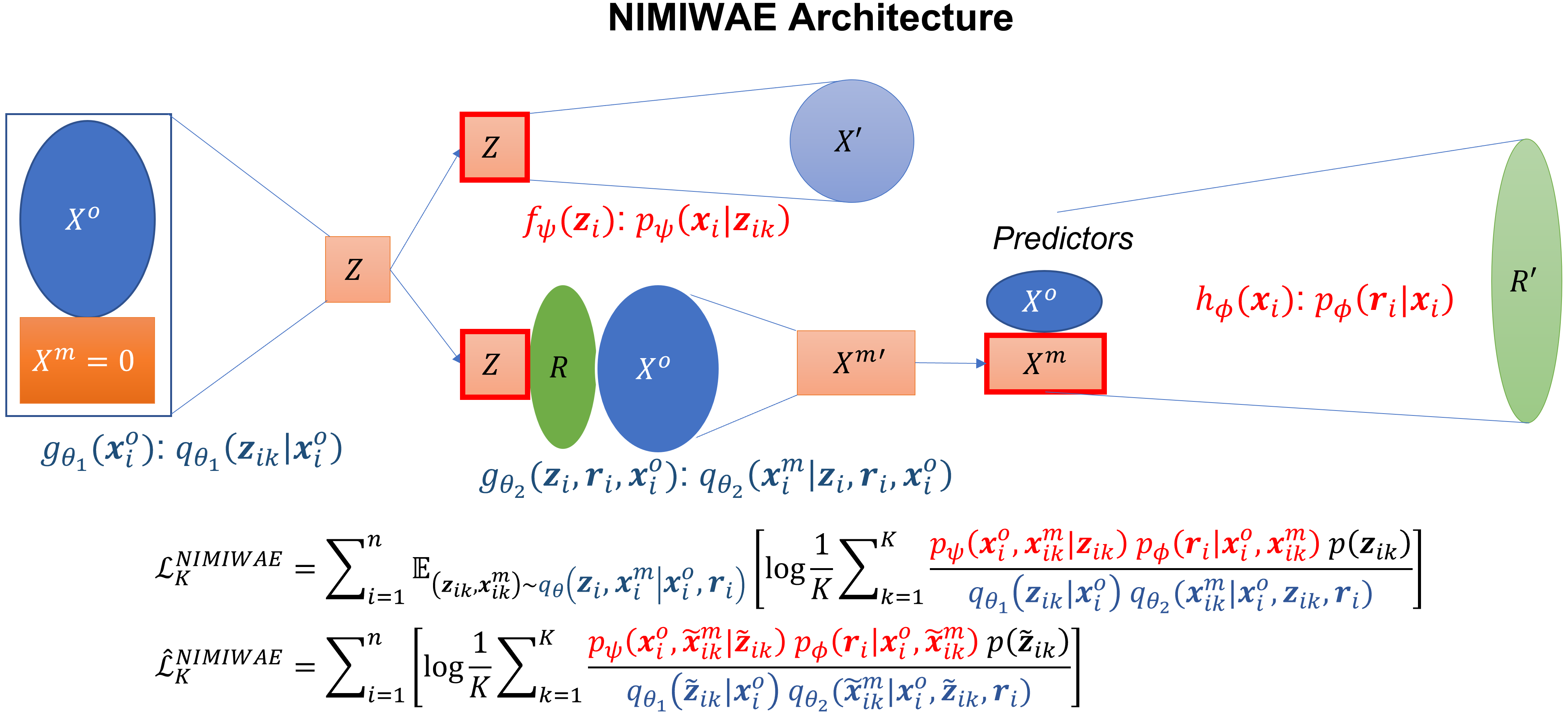} 
\end{center}
\caption{Architecture of proposed NIMIWAE method. Dark colored nodes ($X^o, R, X^m=0$) represent deterministic values, lightly colored nodes ($Z, X^{m\prime}$) represent learned distributional parameters, and outlined (in red) nodes represent sampled values. Orange cells correspond to latent variables $\mathbf{Z}$ and $\mathbf{X}^m$. $\mathbf{Z}_1,\ldots, \mathbf{Z}_K$ and $\mathbf{X}_{1}^m,\ldots,\mathbf{X}_{K}^m$ are sampled from their respective variational posteriors $q_{\theta_1}(\mathbf{Z}|\mathbf{X}^o)$ and $q_{\theta_2}(\mathbf{X}^m|\mathbf{Z},\mathbf{R},\mathbf{X}^o)$. Below is the NIMIWAE bound ($\mathcal{L}_{K}^{NIMIWAE}$) and the estimate of the NIMIWAE bound ($\hat{\mathcal{L}}_{K}^{NIMIWAE}$), which is optimized via stochastic gradient descent.}
\label{fig:NIMIWAEarchitecture}
\end{figure}

The quantity $h_\phi(\mathbf{x}_i)$, which we call the ``missingness network'', outputs the distributional parameters to the missingness model $p_\phi(\mathbf{r}_i|\mathbf{x}_i^o,\tilde{\mathbf{x}}_{ik}^m)$ specifically given in Eq. (\ref{eqn:NIMIWAEbound2}). Furthermore, by omitting this network and its contribution to the NIMIWAE bound altogether, one can attain an ignorably-missing version of the NIMIWAE method (IMIWAE), which would be more suitable for use when assuming MCAR and MAR missingness.  We explore the empirical performance of each of these models under misspecification of the missingness mechanism in Section \ref{sec:examples2}.  An illustration of  $h_\phi(\mathbf{x}_i)$ is given in Figure \ref{fig:NIMIWAEarchitecture}.  


Historically, the set of features for the $p_{miss}$ logistic regression models from Eq. (\ref{eqn:logistic}) need to be carefully pre-specified, usually based upon prior information \citep{Little2002}. Prior work has shown that overparameterization of the missingness model can lead to identifiability issues and divergence in EM-based maximization procedures \citep{Ibrahim2009}.  Our proposed method allows users to similarly pre-specify a subset of features in the missingness network. When such information is not available, it is unclear how such overparameterization will affect model fitting and performance in this setting.  Therefore, in Section \ref{sec:examples2}, we evaluate empirically the use of all $p$ features in the missingness network when assuming MNAR missingness using statistical simulation.

Additional details regarding the NIMIWAE algorithm are outlined in Web Appendix A.



\subsubsection{Multiple Imputation} \label{sec:MI}

Following training, NIMIWAE can provide point estimates for $\E[\mathbf{x}_i^m|\mathbf{x}_i^o,\mathbf{r}_i]$, defined as the expected value of the missing features given the observed data and the mask for the $i^{th}$ observation under MNAR.  The same NIMIWAE model can also perform multiple imputation of the incomplete training dataset to facilitate inference in downstream statistical models. We first note that
\begin{align*}
\E[\mathbf{x}_i^m|\mathbf{x}_i^o, \mathbf{r}_i] &= \int \mathbf{x}_i^m p_{\psi,\phi}(\mathbf{x}_i^m|\mathbf{x}_i^o,\mathbf{r}_i) d\mathbf{x}_i^m \\
&= \iint \mathbf{x}_i^m p_{\psi,\phi}(\mathbf{x}_i^m,\mathbf{z}_i|\mathbf{x}_i^o,\mathbf{r}_i) d\mathbf{z}_i d\mathbf{x}_i^m \\
&= \iint \mathbf{x}_i^m \dfrac{p_{\psi,\phi}(\mathbf{x}_i^m,\mathbf{x}_i^o,\mathbf{z}_i,\mathbf{r}_i)}{p_{\psi,\phi}(\mathbf{x}_i^o,\mathbf{r}_i)} d\mathbf{z}_id\mathbf{x}_i^m \\
&= \iint \mathbf{x}_i^m \dfrac{p_{\psi}(\mathbf{x}_i|\mathbf{z}_{i})p(\mathbf{z}_{i})p_{\phi}(\mathbf{r}_i|\mathbf{x}_i^o,\mathbf{x}_{i}^m)}{p_{\psi,\phi}(\mathbf{x}_i^o,\mathbf{r}_i)} d\mathbf{z}_id\mathbf{x}_i^m.
\end{align*}
Then, we may estimate this integral by self-normalized importance sampling. We utilize the proposal density $q_{\theta_1}(\mathbf{z}_i|\mathbf{x}_i^o) q_{\theta_2}(\mathbf{x}_i^m|\mathbf{z}_i,\mathbf{x}_i^o,\mathbf{r}_i)$, and consider $p_{\psi,\phi}(\mathbf{x}_i^o,\mathbf{r}_i)$ as some unknown constant. Then, we define the quantities 
$$w_{ik} = \dfrac{s_{ik}}{s_{i1}+\ldots+s_{iK}} \textrm{, and } s_{ik}=\dfrac{p_{\psi}(\mathbf{x}_i|\tilde{\mathbf{z}}_{ik})p(\tilde{\mathbf{z}}_{ik})p_{\phi}(\mathbf{r}_i|\mathbf{x}_i^o,\tilde{\mathbf{x}}_{ik}^m)}{q_{\theta_1}(\tilde{\mathbf{z}}_{ik}|\mathbf{x}_i^o) q_{\theta_2}(\tilde{\mathbf{x}}_{ik}^m|\tilde{\mathbf{z}}_{ik},\mathbf{x}_i^o,\mathbf{r}_i)}$$
for $k=1,\ldots,K$, with $1$ sample drawn from the variational posterior of each latent variable $\mathbf{z}_{ik}$ and $\mathbf{x}_{ik}^m$ to compute $s_{ik}$, where $w_{ik}$ is defined as standardized ``importance weights'' \citep{Mattei2019}. Using these weights we may estimate $\E[\mathbf{x}_i^m|\mathbf{x}_i^o,\mathbf{r}_i] \approx \sum_{k=1}^{K} w_{ik}\tilde{\mathbf{x}}_{ik}^m$. Then, the process can be repeated for each observation $i=1,\ldots,n$.

In the MCAR or MAR case, one can similarly estimate $\E[\mathbf{x}_i^m|\mathbf{x}_i^o]$ using the fitted IMIWAE model. By following a similar derivation using the proposal density $q_{\theta_1}(\mathbf{z}_i|\mathbf{x}_i^o)q_{\theta_2}(\mathbf{x}_i^m|\mathbf{z}_i,\mathbf{x}_i^o)$, we obtain the same approximation $\E[\mathbf{x}_i^m|\mathbf{x}_i^o] \approx \sum_{k=1}^K w_{ik}\tilde{\mathbf{x}}_{ik}^m$, with $w_{ik}$ defined as before, but with a slightly different form for $s_{ik}$: $$ s_{ik}=\dfrac{p_{\psi}(\mathbf{x}_i|\tilde{\mathbf{z}}_{ik})p(\tilde{\mathbf{z}}_{ik})}{q_{\theta_1}(\tilde{\mathbf{z}}_{ik}|\mathbf{x}_i^o) q_{\theta_2}(\tilde{\mathbf{x}}_{ik}^m|\tilde{\mathbf{z}}_{ik},\mathbf{x}_i^o)}.$$

Given these weights $w_{ik}$, we may now also produce $Q$ multiply-imputed datasets using the sampling importance resampling (SIR) algorithm \citep{Smith1992}. We first construct a set of $K$ candidate draws and corresponding weights using the procedure described above (default $K = 10\times Q $ draws), and then perform a weighted resample of size $Q$ with replacement from this set of draws to obtain approximate draws from $p_\psi(\mathbf{x}_i^m|\mathbf{x}_i^o,\mathbf{r}_i)$, for each observation $i=1,\ldots,n$. We may then use techniques such as ``Rubin's rules'' \citep{Rubin2004} to pool the estimates obtained from a candidate regression model fit on each imputed dataset, and obtain pooled point estimates and standard errors that account for the uncertainty due to the imputation. In our analyses, we used NIMIWAE to construct $Q=50$ multiply-imputed datasets by drawing $K=500$ times from $q_{\theta_2}(\mathbf{x}_i^m|\mathbf{z}_i,\mathbf{x}_i^o,\mathbf{r}_i)$ for each $i=1,\ldots,n$ after the model was trained. 





\section{Numerical Results} \label{sec:examples2}
\subsection{Simulated Data} \label{sec:simdata2}


We utilize statistical simulation to evaluate the imputation performance of our proposed NIMIWAE and IMIWAE methods and under the assumption of MCAR, MAR, and MNAR missingness.  We note that for each these simulations, we utilize all $p$ features in NIMIWAE's missingness network. We also compared this performance to state-of-the-art missing data methods in machine learning that claim to handle ignorable missingness patterns: HIVAE \citep{Nazabal2018}, VAEAC \citep{Ivanov2019}, MIWAE \citep{Mattei2019}, in addition to the popular MICE method \citep{VanBuuren2011} and a na\"ive mean imputation method. We also included MissForest \citep{Stekhoven2011} in analyses where the model could be fit in a CPU with 32 GB of memory. For all simulations, we divided the full data into training and validation sets with ratio 8:2. For methods that require hyperparameter tuning, each method was trained on the training set for a given set of hyperparameters, the performance of this model was then evaluated on the validation set, and finally the training set was imputed using the model pertaining to the optimal set of hyperparameters (based up on validation set performnance). For methods that required no hyperparameter tuning, we directly imputed just the training set, for consistency across all methods. In Sections \ref{sec:simsetup2}-\ref{sec:simres2}, we evaluate performance on fully synthetic data. Then, in Section \ref{sec:res_UCI}, we simulate missingness into existing datasets from the UCI machine learning repository to preserve non-linearity and interactions between features previously observed.  The simulation setup and performance criteria are described in the subsequent sections. 


\subsubsection{Simulation Setup} \label{sec:simsetup2}
We first evaluate the performance of each method on completely synthetic data.  Here we assume $\mathbf{X}$ is generated such that $\mathbf{X} = \mathbf{Z}\mathbf{W} + \mathbf{B}$, where $\mathbf{W}$ and $\mathbf{B}$ and are matrices of dimensions $d \times p$ and $n \times p$, respectively, $\mathbf{Z}\sim N_{d}(\mathbf{0},\mathbf{I})$, $W_{lj} \sim N(0,0.5)$, and $B_{ij} \sim N(0,1)$ for $i = 1,\ldots,n$, $j = 1,\ldots,p$, and $l = 1,\ldots,d$. 

We then simulate the missingness mask matrix $\mathbf{R}$ such that 30\% of features are partially observed, and 50\% of the observations for each of these features are missing. We generate $r_{ij}$ from the Bernoulli distribution with probability equal to $p(r_{ij_m}=1|\mathbf{x}_i,\boldsymbol{\phi})$, such that $\text{logit} [p(r_{ij_m}=1|\mathbf{x}_i,\boldsymbol{\phi})]=\phi_0+\boldsymbol{\phi}_1\mathbf{x}_i^o+\boldsymbol{\phi}_2\mathbf{x}_i^m$.  Here, we assume that  $j_m=1,\ldots,p_{miss}$ index the missing features, $\boldsymbol{\phi}_1 = \{\phi_{11},\ldots,\phi_{1,p_{obs}}\}$ are the coefficients pertaining to the fully observed features, and $\boldsymbol{\phi}_2 = \{\phi_{21},\ldots,\phi_{2,p_{miss}}\}$ are those pertaining to the partially observed features, and $p_{obs}$ and $p_{miss}$ are the total number of features that are fully and partially observed, respectively.  We set $p_{miss} = \lfloor 0.3*p \rfloor$ and $p_{obs}=p-p_{miss}$. We drew nonzero values of $\boldsymbol{\phi}_1$ and $\boldsymbol{\phi}_2$ from the log-normal distribution with mean $\mu_\phi=5$, with $\log$ standard deviation $\sigma_\phi=0.2$.  

To evaluate the impact of the misspecification of the missingness mechanism on model performance, $r_{ij_m}$ was simulated under each mechanism as follows: (1) MCAR: $\{\boldsymbol{\phi}_1,\boldsymbol{\phi}_2\}=0$, (2) MAR: Same as MCAR except $\phi_{1j_o} \neq 0$ for one randomly selected completely-observed feature $j_o$ where $j_o = p_{miss}+1,\ldots,p$, and  (3) MNAR: Same as MCAR except $\phi_{2j_m} \neq 0$. In this way, for each MAR or MNAR feature, the missingness is dependent on just one feature. In each case, we used $\phi_0$ to adjust the overall expected rate of missingness of each feature to approximately $50\%$. 

Lastly, we simulated a binary response variable assuming Pr$(\mathbf{y}=1|\mathbf{X})=Sigmoid(\beta_0 + \mathbf{X}\boldsymbol{\beta})$, where $\mathbf{y}$ is a binary response variable, $\boldsymbol{\beta}=\{\beta_1,\ldots,\beta_p\}$ are the set of regression coefficients, and $\beta_0$ is the intercept. In many applications, it is of interest to use the features in $\boldsymbol{X}$ to predict some outcome variable $\mathbf{y}$ when $\boldsymbol{X}$ is only partially observed.  Therefore, the ability accurately estimate $\boldsymbol{\beta}$ is also of importance in the presence of missingness. Multiple imputation methods like MICE can be used to perform coefficient estimation by using Rubin's rules to pool the coefficient estimates from logistic regression models fitted on each individual multiply-imputed dataset. We similarly perform coefficient estimation using multiply-imputed datasets from the SIR algorithm within NIMIWAE, allowing for direct comparisons in coefficient estimation with MICE in estimating $\boldsymbol{\beta}$. 

We vary $n$, $p$, and $d$ such that $n=\{10,000, 100,000\}$, $p=\{25,100\}$ features, and $d=\{2,8\}$. We simulated 5 datasets per simulation condition, spanning various missingness mechanisms and values for $\{n,p,d\}$. We fix the values of $\boldsymbol{\beta}$ at $0.25$ for each feature, and adjusted $\beta_0$ to ensure equal proportions for the two binary classes in $\mathbf{Y}$. We measured imputation performance by calculating the average L1 distance between true and imputed masked values in $\mathbf{X}$. Letting $\hat{\mathbf{X}}^m$ denote the imputed masked values of the true $\mathbf{X}^m$ values of the missing entries, we denote the average L1 distance is simply  $\frac{\mid \hat{\mathbf{X}}^m - \mathbf{X}^m \mid}{N_{miss}},$  where $N_{miss}$ is the total number of missing entries in the dataset. To assess the ability of multiple imputation methods in performing coefficient estimation, we reported the percent bias (PB) of these pooled estimates compared to the truth, averaged across the $p$ features, i.e.  $PB = \frac{1}{p}\sum_{j=1}^p \frac{|\beta_j-\hat{\beta}_j|}{|\beta_j|}.$ 

\begin{figure}
\begin{center}
\includegraphics[width=175mm]{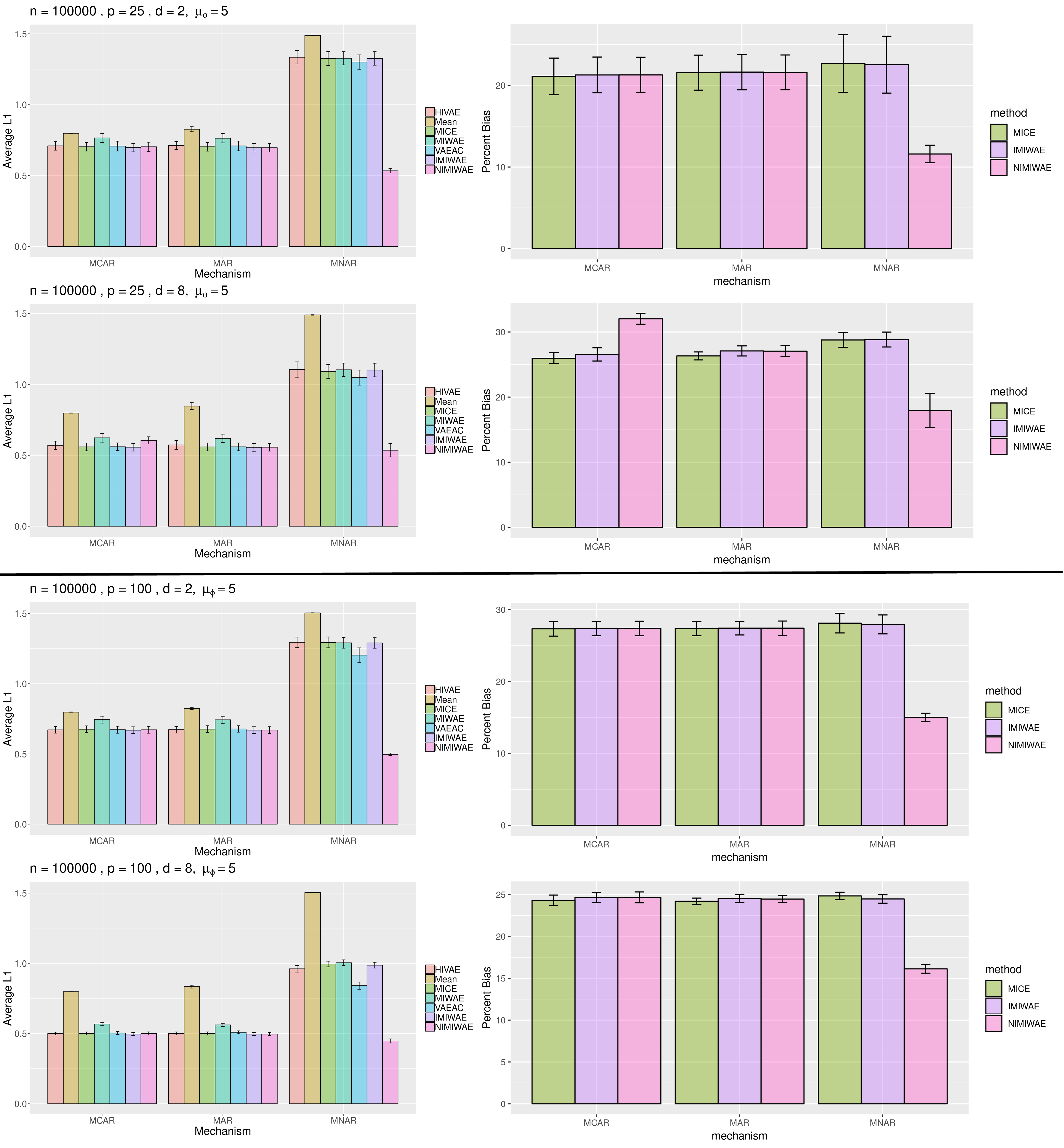} 
\end{center}
\caption{\small Average L1 distance between true and imputed values for missing entries (left) and percent bias of pooled coefficient estimates (right) for $p=25$ (top 4) and $p=100$ (bottom 4) features, stratified by $d$. NIMIWAE outperforms all methods in imputing MNAR missing values, while performing comparably to other methods in imputing MCAR and MAR values. Here, $n=100,000$, $\mu_\phi=5$, and error bars show the variability of each metric across 5 reps. Weights and biases were initialized by using the default semi-orthogonal matrix method.}
\label{fig:simsN100K}
\end{figure}

\subsubsection{Simulation Results} \label{sec:simres2}
Figure \ref{fig:simsN100K} shows the results pertaining to $n=100,000$.  Error bars represented the variability in performance across 5 simulations. Despite the overparameterized missingness model, NIMIWAE consistently yields improved imputation performance compared to other methods under MNAR missingness, while yielding an average L1 that is comparable to other methods in the MCAR and MAR cases, under these conditions assuming large sample sizes. This can be exceptionally useful for many real data applications, where the true covariates of the missingness model may not be known \textit{a priori}. In estimating $\boldsymbol{\beta}$, we see that NIMIWAE yields a lower average percent bias under MNAR missingness, while MICE, IMIWAE, and NIMIWAE perform similarly under MCAR or MAR missingness. 


\begin{figure}
\begin{center}
\includegraphics[width=175mm]{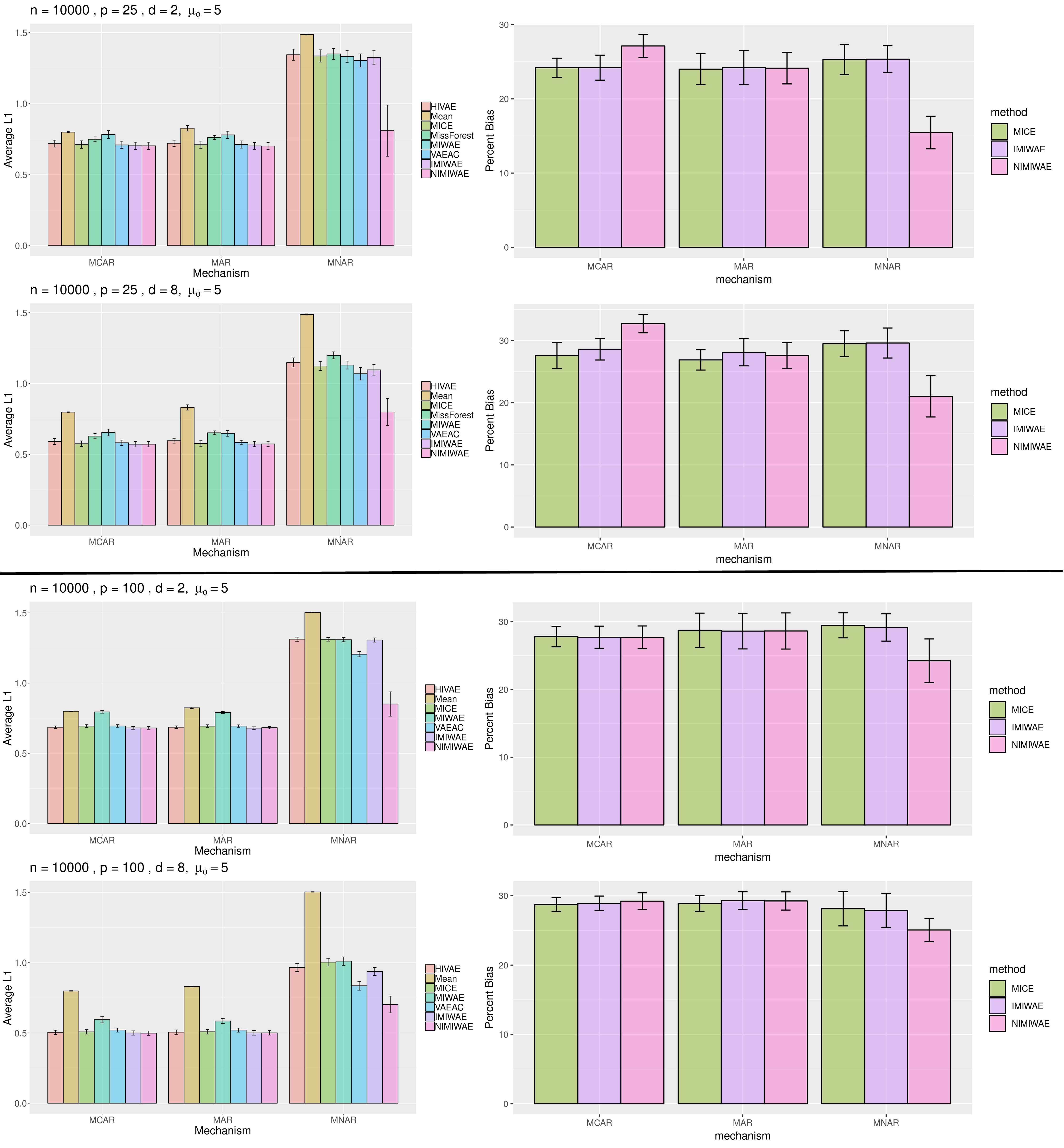} 
\end{center}
\caption{\small Average L1 distance between true and imputed values for missing entries (left) and percent bias of pooled coefficient estimates (right) for $p=25$ (top 4) and $p=100$ (bottom 4) features, stratified by $d$. NIMIWAE outperforms all methods in imputing MNAR missing values, while performing comparably to other methods in imputing MCAR and MAR values. Here, $n=10,000$, $\mu_\phi=5$, and error bars show the variability of each metric across 5 reps. Weights and biases were initialized by using the alternative method, as described in Section \ref{sec:VAENonignorable}}
\label{fig:simsN10K}
\end{figure}

Figure \ref{fig:simsN10K} shows the results pertaining to $n=10,000$ utilizing the alternative initialization method described in Section \ref{sec:VAENonignorable}, due to the smaller sample size in this setting. We see that NIMIWAE performs best in imputing MNAR values, and in estimating the coefficients under MNAR missingness, and still performs comparably well to other methods in the MCAR and MAR cases. The results of the analyses on $n=10,000$ simulations with default initialization can be found in Web Appendix B. We found that the default method initialized weights for missing features too small for the network to recover in the MNAR case, under the lower sample size and high dimensionality setting. Based upon these results, we recommend the alternative weight initialization for smaller sample sizes ($n \leq 10,000$) with large dimensionality ($p \geq 100$), while the default initialization may suffice when the sample size is large, or if the dimensionality of the data is smaller. Alternatively, one may greatly improve imputation performance by narrowing down the features that are input into the missingness network using some prior knowledge or assumptions on the mechanism of missingness. In practical applications, it is unknown which mechanism may truly underly the data, and, as usual, sensitivity analyses, where one varies the assumed mechanism, is still important.  For example, imputing data via IMIWAE may be helpful and be more efficient under MCAR or MAR than via NIMIWAE. 

Overall, these simulations confirm that existing methods can impute MCAR and MAR missingness with a reasonable degree of accuracy, but they break down under MNAR missingness. Our NIMIWAE method is able to adequately impute MNAR missing values, while still imputing MCAR and MAR missing values with a comparable degree of accuracy as existing methods geared specifically towards the ignorable mechanisms of missingness. Additionally, we note that MissForest was able to be run to the $n=10,000$ and $p=25$ simulation case only due to memory constraints, and yielded very poor imputation performance in the MNAR case, like the other ignorably-missing methods.  We also show empirically that using all $p$ features in NIMIWAE's missingness network can yield reasonable performance, especially when the number of samples is large.  We postulate that this may be due to the fact that neural networks have been shown to generalize well despite severe overparameterization  \citep{Poggio2020}. Still, the specification of a smaller model that is closer to the truth may improve the accuracy of imputations, especially under smaller sample sizes \citep{Du2021}.

\subsubsection{UCI Machine Learning Datasets} \label{sec:res_UCI}
Next, we analyzed how accurately these methods can impute missing values that may be present in real data. Since we are unable to obtain the true missing values in real datasets, we selected 6 different completely-observed datasets from the UCI Machine Learning Repository, and simulated missingness according to each mechanism as in the fully synthetic datasets. Here, we masked half of the features, such that $p_{miss}=floor(0.5*p)$. Additional details on these datasets and how they may be obtained can be found in Web Appendix B, and the values of the hyperparameters that were tuned can be found in Web Appendix C.

\begin{table}[ht]
\centering
\begingroup\fontsize{11pt}{12.5pt}\selectfont
\begin{tabular}{llrrrrrrrr}
  \hline
{\footnotesize Dataset} & {\footnotesize  } & {\footnotesize HIVAE} & {\footnotesize Mean} & {\footnotesize MICE} & {\footnotesize MissForest} & {\footnotesize MIWAE} & {\footnotesize VAEAC} & {\footnotesize NIMIWAE} & {\footnotesize IMIWAE} \\
  \hline
banknote & \textit{ MCAR } & 1.62 & 2.00 & 1.63 & \textcolor{red}{0.95} & 1.37 & 1.32 & 1.49 & 1.19 \\
{\footnotesize $n=1372$}   & \textit{ MAR } & 2.00 & 2.27 & 2.43 & 1.88 & 2.90 & \textcolor{red}{1.76} & 1.89 & 1.96 \\
   & \textit{ MNAR } & 3.39 & 3.92 & 3.78 & 3.18 & 3.10 & 3.46 & \textcolor{red}{1.46} & 3.30 \\
   \rowcolor[gray]{0.95}concrete & \textit{ MCAR } & 47.54 & 51.28 & 29.97 & \textcolor{red}{25.57} & 40.53 & 33.03 & 42.12 & 33.69 \\
   \rowcolor[gray]{0.95}{\footnotesize $n=1030$} & \textit{ MAR } & 67.37 & 60.00 & \textcolor{red}{44.77} & 53.19 & 66.35 & 61.09 & 55.82 & 57.56 \\
   \rowcolor[gray]{0.95} & \textit{ MNAR } & 59.48 & 95.85 & 68.24 & 79.87 & 76.79 & 70.12 & \textcolor{red}{47.46} & 74.44 \\
  hepmass & \textit{ MCAR } & 0.75 & 0.82 & 0.74 &  NA & 0.80 & \textcolor{red}{0.68} & 0.78 & 0.69 \\
  {\footnotesize $n=525,123$} & \textit{ MAR } & 0.76 & 0.84 & 0.75 &  NA & 0.84 & 0.72 & 0.72 & \textcolor{red}{0.71} \\
   & \textit{ MNAR } & 1.41 & 1.54 & 1.40 & NA & 1.36 & 1.30 & \textcolor{red}{0.99} & 1.36 \\
   \rowcolor[gray]{0.95}power & \textit{ MCAR } & 0.54 & 0.66 & 0.50 & NA & 0.59 & \textcolor{red}{0.48} & 0.56 & 0.49 \\
   \rowcolor[gray]{0.95}{\footnotesize $n=1,000,000$} & \textit{ MAR } & 0.61 & 0.75 & 0.56 & NA & 0.67 & \textcolor{red}{0.54} & 0.78 & 0.57 \\
   \rowcolor[gray]{0.95} & \textit{ MNAR } & 0.75 & 1.14 & 0.84 & NA & 0.86 & 0.79 & \textcolor{red}{0.73} & 0.79 \\
  red & \textit{ MCAR } & 1.15 & 1.64 & 1.04 & \textcolor{red}{0.86} & 1.10 & 1.10 & 1.08 & 0.98 \\
  {\footnotesize $n=1599$} & \textit{ MAR } & 1.23 & 1.67 & 1.16 & \textcolor{red}{0.98} & 1.30 & 1.29 & 1.09 & 1.06 \\
   & \textit{ MNAR } & 2.12 & 3.24 & 2.46 & 2.06 & 1.87 & 2.73 & \textcolor{red}{0.90} & 1.74 \\
   \rowcolor[gray]{0.95}white & \textit{ MCAR } & 2.14 & 2.61 & 1.97 & \textcolor{red}{1.60} & 2.18 & 1.93 & 2.16 & 1.88 \\
   \rowcolor[gray]{0.95}{\footnotesize $n=4898$} & \textit{ MAR } & 2.28 & 2.63 & 1.99 & \textcolor{red}{1.69} & 2.15 & 2.11 & 1.89 & 1.89 \\
   \rowcolor[gray]{0.95} & \textit{ MNAR } & 4.42 & 5.36 & 4.21 & 4.27 & 4.46 & 4.10 & \textcolor{red}{3.47} & 4.70 \\
   \hline
\end{tabular}
\endgroup
\caption{Average L1 distance between true and imputed missing values in various datasets, under different mechanisms of simulated missingness. Best imputation performance (lowest average L1) in each row is highlighted in red. Proportion of missing entries was fixed at 50\% per feature, with 50\% of the features containing missingness. We see that NIMIWAE consistently performs best in imputing MNAR missingness, while performance of the ``Ignorable'' IMIWAE model is comparable to other methods under MCAR and MAR. Although MissForest claims superiority in MCAR and some MAR cases in the smaller datasets, it was not scalable to larger datasets like hepmass and power.}
\label{tab:UCI}
\end{table}

Table \ref{tab:UCI} shows the average L1 distance between true and imputed missing values for each of the 6 UCI datasets, with each simulated mechanism of missingness. We see that NIMIWAE again performs best across all methods in accurately imputing MNAR missing values. Overall, MissForest performed best in imputing MCAR missing values in the smaller UCI datasets, but this method was too memory-consuming to be trained on the significantly larger \textit{hepmass} and \textit{power} datasets. As in the simulated data, mean imputation is consistently one of the least accurate methods of imputation. Also, whereas MICE performed very well under MCAR and MAR in the simulated data, deep learning methods like VAEAC and IMIWAE consistently yield more accurate imputations here. This may be due to the fact that MICE uses a fully-conditional linear model for imputation, it may therefore perform suboptimally when the true relationships between features are non-linear. Also, we see that NIMIWAE imputes values slightly less accurately than IMIWAE when the missingness is MCAR or MAR, since in these cases NIMIWAE explicitly estimates the missingness network when it is not necessary.

\subsection{Physionet 2012 Challenge Dataset} \label{sec:physionet}
Finally, we analyzed the Physionet 2012 Challenge data using each of the compared methods. This data consisted of 114 features of 12,000 patients after pre-processing, and the details of the acquisition and preparation of the data can be found in Web Appendix C. We performed a qualitative analysis of each imputed dataset, highlighting differences between the results of downstream regression models fitted on mortality, assuming non-ignorable (NIMIWAE) and ignorable missingness  \citep{Ibrahim2005,Ibrahim2009}. This is because, in contrast to our simulations, the true values of the missing entries are not available to directly assess imputation performance. Additionally, the missingness mechanism itself is generally not ``testable'' by the observed data in practice \citep{Ibrahim1999}.  Here, we used the alternative initialization scheme of NIMIWAE, due to the limited sample size and large dimension of the data. Additionally, we added ``supervised'' versions of NIMIWAE, IMIWAE, and MICE in our comparisons, where the response of interest (mortality) is included with the features of the data during training and imputation, such that the response variable may inform the multiple imputations. After model fitting, we performed Rubin's rules to pool the estimates of the fitted models on the multiply-imputed datasets, and compute standard errors across imputations.

\begin{sidewaysfigure}
\centering
\includegraphics[width=240mm]{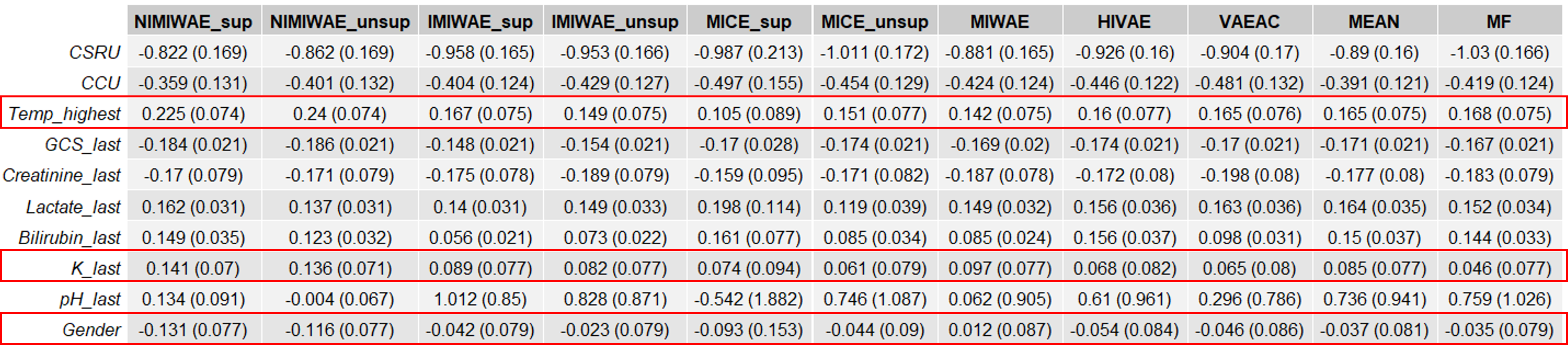}
\caption{\small Table of coefficient estimates (and standard errors) of covariates with the top 10 magnitudes of estimates via $NIMIWAE_{sup}$, from fitting a logistic regression model with imputed datasets from each method. Results from multiple imputation methods NIMIWAE, IMIWAE, and MICE (first 6 columns) are based on 50 multiply imputed datasets, and reflect pooled coefficient estimates and standard errors using Rubin's rules. For fair comparison, we also included results from single imputation methods (last 5 columns). Here, IMIWAE is the ignorable version of NIMIWAE.}
\label{tab:Physionet_LR}
\end{sidewaysfigure}

Based on the imputed datasets from these methods, we fit a logistic regression model with post-baseline mortality as the binary response, with the $p=114$ baseline features of this dataset as the covariates. We report details of the covariates with the top 10 largest effects on mortality when using the multiply imputed datasets from the supervised non-ignorably missing NIMIWAE model in Table \ref{tab:Physionet_LR}. We found some significant differences in the results of the logistic regression based on the different imputed datasets. For example, we found that $NIMIWAE_{sup}$ uncovered a stronger effect of the variables $Temp\_highest$, $K\_last$ and $Gender$ compared to other methods. These factors have been studied for their association with mortality in ICU patients. Specifically, gender has been studied for having a potentially significant effect on mortality in critically-ill patients \citep{Mahmood2012,Larsson2019}. Additionally, body temperature \citep{SchellChaple2015} and irregular potassium levels \citep{Tongyoo2018} have both been known to be associated with an increased risk of in-hospital mortality.


\begin{figure}
\begin{center}
\includegraphics[width=175mm]{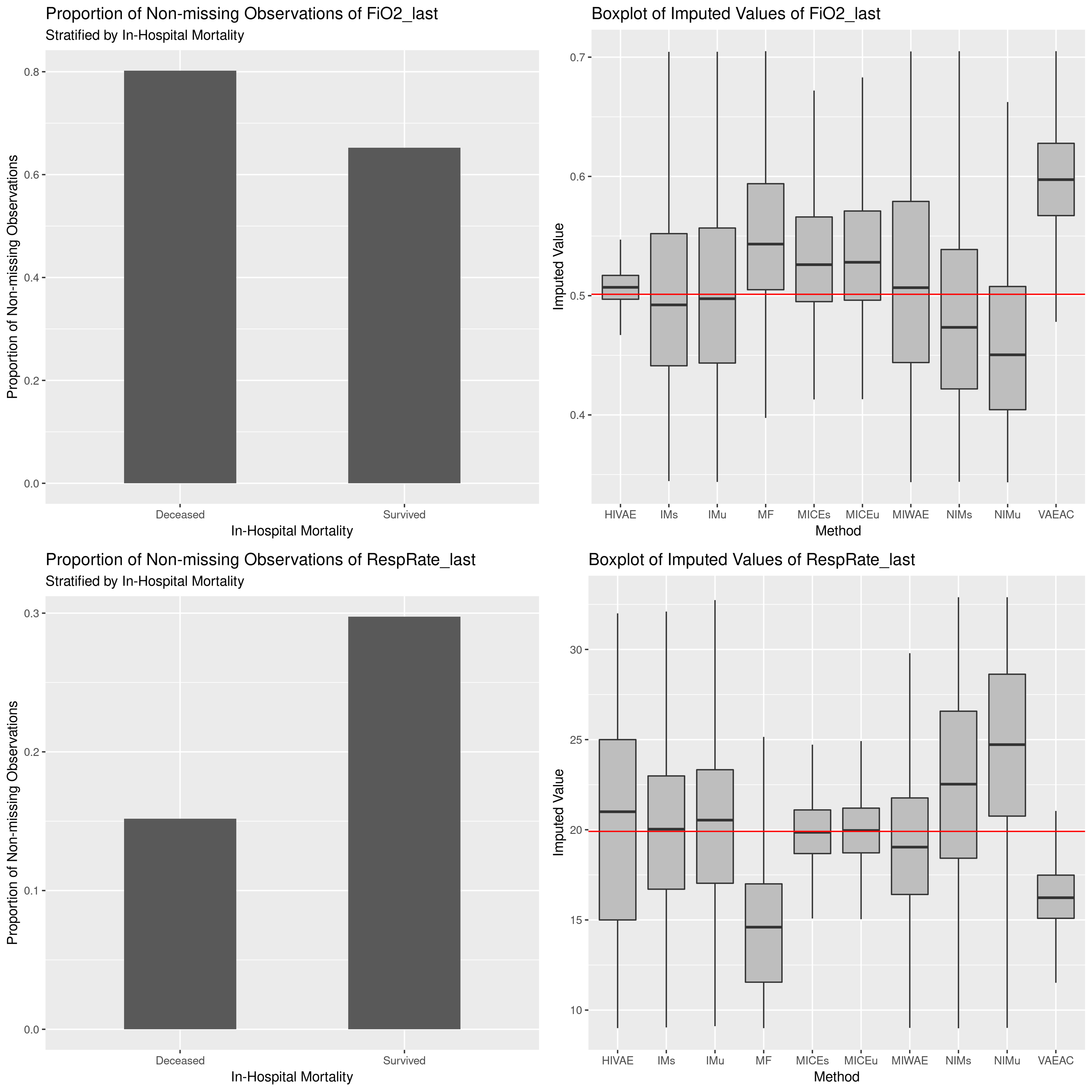} 
\end{center}
\caption{\small (left) Proportion of non-missing observations of $FiO2\_last$ (top) and $RespRate\_last$ (bottom) in surviving and deceased ICU patients, and (right) imputed values of non-missing entries by HIVAE, supervised and unsupervised versions of the ignorable NIMIWAE (IMs, IMu), MissForest (MF), supervised and unsupervised versions of MICE (MICEs, MICEu), MIWAE, supervised and unsupervised versions of NIMIWAE (NIMs, NIMu), and VAEAC. The mean of the observed values is given by the red horizontal line.}
\label{fig:Physionet_imputation}
\end{figure}

Figure \ref{fig:Physionet_imputation} shows the imputed values of two variables that had significantly different missingness rates in surviving vs deceased patients, $FiO2\_last$ and $RespRate\_last$. Imputed values from each of the methods are shown in the boxplots on the right, and the rates of missingness in the deceased vs. surviving patients for each variable are shown on the left. We found that a larger proportion of entries of $FiO2\_last$ and a smaller proportion of entries of $RespRate\_last$ were observed in the surviving patients than in the deceased patients. Of the methods we used to impute values of $FiO2\_last$, we found that $NIMIWAE_{sup}$ and $NIMIWAE_{unsup}$ imputed values were generally smaller than those from other methods. We also found that the supervised and unsupervised $IMIWAE$ models yielded similar values to $HIVAE$, $MIWAE$, and $MICE$. These are all ignorably-missing methods, and may not impute accurate values when the missingness is MNAR. MissForest and VAEAC imputed values of FiO2 that were significantly higher than the observed mean, and values of RespRate that were significantly lower than the observed mean. Some studies have shown that abnormally high or low respiratory rates may be associated with higher mortality in critically-ill patients \citep{Straus2014,Ljunggren2016}, suggesting that the missing values of $RespRate\_last$, which were more prevalent in deceased patients, may have truly been further away from the normal range of 12-20. Additionally, \citet{Esteban2002} found that higher levels of administered FiO2 were associated linearly with higher mortality. Thus, the missing values of $FiO2\_last$, which were more prevalent in surviving patients, may have been lower than the observed values. This suggests that the mechanism of missingness of $FiO2\_last$ and $RespRate\_last$ may be non-ignorable, since $NIMIWAE$ imputed more realistic values for these variables according to the mortality rates of patients with missingness in these variables.

\section{Discussion} \label{sec:disc}

In this paper we introduce NIMIWAE, one of the first methods to handle up to MNAR patterns of missingness in the VAE/IWAE class of methods, to address complex patterns of missingess observed in the Physionet EHR data. Using statistical simulations, we show that NIMIWAE performs well in imputing missing features under MNAR, and has reasonable performance under the MCAR and MAR settings. Performance in imputing MCAR and MAR missingness can be further improved in NIMIWAE by using the ignorable version of this model (IMIWAE), where we omit the missingness network. We also found that the results of the analysis on the Physionet data are highly dependent on the choice of missingness model, which specifies the assumption of the missingness mechanism. However, NIMIWAE is able to impute missing values well in simulations regardless of the underlying missingness mechanism, flexibly modelling the mechanism using a deeply-learned neural network. The NIMIWAE-imputed dataset resulted in more realistic imputed values with respect to what we may expect in the Physionet Challenge patients, since NIMIWAE takes into account possible non-ignorable missingness in the data. Additionally, the NIMIWAE architecture learns a lower-dimensional representation of the data, which can be used for tasks such as patient subgroup identification or visualization of data.  

Learning algorithms that can be applied to EHR data like the Physionet Challenge dataset can be valuable tools that clinicians can use to aid decisions in hospital settings and understand patterns within these health records. For example, properly handling missingness in EHRs when imputing the missing entries can improve the performance of prediction algorithms that can assess risk of death or other outcomes of interest, like disease. Informative missingness is a common problem in analyzing EHR data, and accounting for such missingness can be helpful in obtaining accurate, unbiased estimates of the true missing values. We note that although we have used our NIMIWAE method primarily to analyze the Physionet 2012 Challenge dataset, it can more generally be applied to settings where one wishes to train a VAE when missingness is present among input features.



\backmatter


\bibliographystyle{biom}
\bibliography{P2}

\begin{thebibliography}{}

\bibitem[\protect\citeauthoryear{Beaulieu-Jones and and}{Beaulieu-Jones and
  and}{2016}]{Beaulieu-Jones2016}
Beaulieu-Jones, B.~K. and and, J. H.~M. (2016).
\newblock {Missing} {Data} {Imputation} {in} {the} {Electronic} {Health}
  {Record} {Using} {Deeply} {Learned} {Autoencoders}.
\newblock In {\em Biocomputing 2017}. {WORLD} {SCIENTIFIC}.

\bibitem[\protect\citeauthoryear{Bishop}{Bishop}{2006}]{Bishop2006}
Bishop, C.~M. (2006).
\newblock {\em Pattern Recognition and Machine Learning}.
\newblock Springer-Verlag New York Inc.

\bibitem[\protect\citeauthoryear{{Burda}, {Grosse}, and
  {Salakhutdinov}}{{Burda} et~al.}{2015}]{Burda2015}
{Burda}, Y., {Grosse}, R., and {Salakhutdinov}, R. (2015).
\newblock {Importance Weighted Autoencoders}.
\newblock {\em arXiv e-prints} page arXiv:1509.00519.

\bibitem[\protect\citeauthoryear{{Cremer}, {Morris}, and {Duvenaud}}{{Cremer}
  et~al.}{2017}]{Cremer2017}
{Cremer}, C., {Morris}, Q., and {Duvenaud}, D. (2017).
\newblock {Reinterpreting Importance-Weighted Autoencoders}.
\newblock {\em arXiv e-prints} page arXiv:1704.02916.

\bibitem[\protect\citeauthoryear{Diggle and Kenward}{Diggle and
  Kenward}{1994}]{Diggle1994}
Diggle, P. and Kenward, M.~G. (1994).
\newblock Informative drop-out in longitudinal data analysis.
\newblock {\em Applied Statistics} {\bf 43,} 49.

\bibitem[\protect\citeauthoryear{{Doersch}}{{Doersch}}{2016}]{Doersch2016}
{Doersch}, C. (2016).
\newblock {Tutorial on Variational Autoencoders}.
\newblock {\em arXiv e-prints} page arXiv:1606.05908.

\bibitem[\protect\citeauthoryear{Du, Enders, Keller, Bradbury, and Karney}{Du
  et~al.}{2021}]{Du2021}
Du, H., Enders, C., Keller, B.~T., Bradbury, T.~N., and Karney, B.~R. (2021).
\newblock A bayesian latent variable selection model for nonignorable
  missingness.
\newblock {\em Multivariate Behavioral Research} pages 1--49.

\bibitem[\protect\citeauthoryear{Esteban}{Esteban}{2002}]{Esteban2002}
Esteban, A. (2002).
\newblock Characteristics and outcomes in adult patients receiving mechanical
  ventilation. a 28-day international study.
\newblock {\em {JAMA}} {\bf 287,} 345.

\bibitem[\protect\citeauthoryear{Gershman and Goodman}{Gershman and
  Goodman}{2014}]{Gershman2014}
Gershman, S.~J. and Goodman, N.~D. (2014).
\newblock Amortized inference in probabilistic reasoning.
\newblock In {\em CogSci}.

\bibitem[\protect\citeauthoryear{Ibrahim}{Ibrahim}{2001}]{Ibrahim2001}
Ibrahim, J.~G. (2001).
\newblock Missing responses in generalised linear mixed models when the missing
  data mechanism is nonignorable.
\newblock {\em Biometrika} {\bf 88,} 551--564.

\bibitem[\protect\citeauthoryear{Ibrahim, Chen, Lipsitz, and Herring}{Ibrahim
  et~al.}{2005}]{Ibrahim2005}
Ibrahim, J.~G., Chen, M.-H., Lipsitz, S.~R., and Herring, A.~H. (2005).
\newblock Missing-data methods for generalized linear models.
\newblock {\em Journal of the American Statistical Association} {\bf 100,}
  332--346.

\bibitem[\protect\citeauthoryear{Ibrahim, Lipsitz, and Chen}{Ibrahim
  et~al.}{1999}]{Ibrahim1999}
Ibrahim, J.~G., Lipsitz, S.~R., and Chen, M.-H. (1999).
\newblock Missing covariates in generalized linear models when the missing data
  mechanism is non-ignorable.
\newblock {\em Journal of the Royal Statistical Society: Series B (Statistical
  Methodology)} {\bf 61,} 173--190.

\bibitem[\protect\citeauthoryear{Ibrahim and Molenberghs}{Ibrahim and
  Molenberghs}{2009}]{Ibrahim2009}
Ibrahim, J.~G. and Molenberghs, G. (2009).
\newblock Missing data methods in longitudinal studies: a review.
\newblock {\em {TEST}} {\bf 18,} 1--43.

\bibitem[\protect\citeauthoryear{Ivanov, Figurnov, and Vetrov}{Ivanov
  et~al.}{2019}]{Ivanov2019}
Ivanov, O., Figurnov, M., and Vetrov, D. (2019).
\newblock Variational autoencoder with arbitrary conditioning.
\newblock In {\em International Conference on Learning Representations}.

\bibitem[\protect\citeauthoryear{{Kingma} and {Ba}}{{Kingma} and
  {Ba}}{2014}]{Kingma2014}
{Kingma}, D.~P. and {Ba}, J. (2014).
\newblock {Adam: A Method for Stochastic Optimization}.
\newblock {\em arXiv e-prints} page arXiv:1412.6980.

\bibitem[\protect\citeauthoryear{{Kingma} and {Welling}}{{Kingma} and
  {Welling}}{2019}]{Kingma2019}
{Kingma}, D.~P. and {Welling}, M. (2019).
\newblock {An Introduction to Variational Autoencoders}.
\newblock {\em arXiv e-prints} page arXiv:1906.02691.

\bibitem[\protect\citeauthoryear{Larsson, Lindström, Eriksson, and
  Oldner}{Larsson et~al.}{2019}]{Larsson2019}
Larsson, E., Lindström, A.-C., Eriksson, M., and Oldner, A. (2019).
\newblock Impact of gender on post- traumatic intensive care and outcomes.
\newblock {\em Scandinavian Journal of Trauma, Resuscitation and Emergency
  Medicine} {\bf 27,}.

\bibitem[\protect\citeauthoryear{LeCun, Bengio, and Hinton}{LeCun
  et~al.}{2015}]{LeCun2015}
LeCun, Y., Bengio, Y., and Hinton, G. (2015).
\newblock Deep learning.
\newblock {\em Nature} {\bf 521,} 436--444.

\bibitem[\protect\citeauthoryear{Li, Akbar, and Oliva}{Li
  et~al.}{2020}]{Li2020}
Li, Y., Akbar, S., and Oliva, J. (2020).
\newblock Acflow: Flow models for arbitrary conditional likelihoods.
\newblock In {\em International Conference on Machine Learning}, pages
  5831--5841. PMLR.

\bibitem[\protect\citeauthoryear{Lim, Jiang, and Yi}{Lim
  et~al.}{2020}]{Lim2020}
Lim, K.-L., Jiang, X., and Yi, C. (2020).
\newblock Deep clustering with variational autoencoder.
\newblock {\em IEEE Signal Processing Letters} {\bf 27,} 231--235.

\bibitem[\protect\citeauthoryear{Little and Rubin}{Little and
  Rubin}{2002}]{Little2002}
Little, R. J.~A. and Rubin, D.~B. (2002).
\newblock {\em Statistical Analysis with Missing Data}.
\newblock John Wiley {\&} Sons, Inc.

\bibitem[\protect\citeauthoryear{Ljunggren, Castr{\'{e}}n, Nordberg, and
  Kurland}{Ljunggren et~al.}{2016}]{Ljunggren2016}
Ljunggren, M., Castr{\'{e}}n, M., Nordberg, M., and Kurland, L. (2016).
\newblock The association between vital signs and mortality in a retrospective
  cohort study of an unselected emergency department population.
\newblock {\em Scandinavian Journal of Trauma, Resuscitation and Emergency
  Medicine} {\bf 24,}.

\bibitem[\protect\citeauthoryear{Lopez, Regier, Cole, Jordan, and Yosef}{Lopez
  et~al.}{2018}]{Lopez2018}
Lopez, R., Regier, J., Cole, M.~B., Jordan, M.~I., and Yosef, N. (2018).
\newblock Deep generative modeling for single-cell transcriptomics.
\newblock {\em Nature Methods} {\bf 15,} 1053--1058.

\bibitem[\protect\citeauthoryear{Mahmood, Eldeirawi, and Wahidi}{Mahmood
  et~al.}{2012}]{Mahmood2012}
Mahmood, K., Eldeirawi, K., and Wahidi, M.~M. (2012).
\newblock Association of gender with outcomes in critically ill patients.
\newblock {\em Critical Care} {\bf 16,} R92.

\bibitem[\protect\citeauthoryear{Mattei and Frellsen}{Mattei and
  Frellsen}{2019}]{Mattei2019}
Mattei, P.-A. and Frellsen, J. (2019).
\newblock {MIWAE}: Deep generative modelling and imputation of incomplete data
  sets.
\newblock In Chaudhuri, K. and Salakhutdinov, R., editors, {\em Proceedings of
  the 36th International Conference on Machine Learning}, volume~97 of {\em
  Proceedings of Machine Learning Research}, pages 4413--4423, Long Beach,
  California, USA. PMLR.

\bibitem[\protect\citeauthoryear{Murphy}{Murphy}{2016}]{Murphy2016}
Murphy, J. (2016).
\newblock An overview of convolutional neural network architectures for deep
  learning.
\newblock {\em Microway Inc} pages 1--22.

\bibitem[\protect\citeauthoryear{{Nazabal}, {Olmos}, {Ghahramani}, and
  {Valera}}{{Nazabal} et~al.}{2018}]{Nazabal2018}
{Nazabal}, A., {Olmos}, P.~M., {Ghahramani}, Z., and {Valera}, I. (2018).
\newblock {Handling Incomplete Heterogeneous Data using VAEs}.
\newblock {\em arXiv e-prints} page arXiv:1807.03653.

\bibitem[\protect\citeauthoryear{{O'Shea}}{{O'Shea}}{2019}]{OShea2019}
{O'Shea}, R. (2019).
\newblock {Interpreting Missing Data Patterns in the ICU}.
\newblock {\em arXiv e-prints} page arXiv:1912.08612.

\bibitem[\protect\citeauthoryear{Poggio, Banburski, and Liao}{Poggio
  et~al.}{2020}]{Poggio2020}
Poggio, T., Banburski, A., and Liao, Q. (2020).
\newblock Theoretical issues in deep networks.
\newblock {\em Proceedings of the National Academy of Sciences} {\bf 117,}
  30039--30045.

\bibitem[\protect\citeauthoryear{Prechelt}{Prechelt}{1998}]{Prechelt1998}
Prechelt, L. (1998).
\newblock Early stopping-but when?
\newblock In {\em Neural Networks: Tricks of the trade}, pages 55--69.
  Springer.

\bibitem[\protect\citeauthoryear{Ross, Wei, and Ohno-Machado}{Ross
  et~al.}{2014}]{Ross2014}
Ross, M.~K., Wei, W., and Ohno-Machado, L. (2014).
\newblock {\textquotedblleft}big data{\textquotedblright} and the electronic
  health record.
\newblock {\em Yearbook of Medical Informatics} {\bf 23,} 97--104.

\bibitem[\protect\citeauthoryear{Rubin}{Rubin}{1976}]{Rubin1976}
Rubin, D.~B. (1976).
\newblock Inference and missing data.
\newblock {\em Biometrika} {\bf 63,} 581--592.

\bibitem[\protect\citeauthoryear{Rubin}{Rubin}{2004}]{Rubin2004}
Rubin, D.~B. (2004).
\newblock {\em Multiple imputation for nonresponse in surveys}, volume~81.
\newblock John Wiley \& Sons.

\bibitem[\protect\citeauthoryear{Saxe, Mcclelland, and Ganguli}{Saxe
  et~al.}{2014}]{Saxe2014}
Saxe, A.~M., Mcclelland, J.~L., and Ganguli, S. (2014).
\newblock Exact solutions to the nonlinear dynamics of learning in deep linear
  neural network.
\newblock In {\em In International Conference on Learning Representations}.

\bibitem[\protect\citeauthoryear{Schell-Chaple, Puntillo, Matthay, and
  and}{Schell-Chaple et~al.}{2015}]{SchellChaple2015}
Schell-Chaple, H.~M., Puntillo, K.~A., Matthay, M.~A., and and, K. D.~L.
  (2015).
\newblock Body temperature and mortality in patients with acute respiratory
  distress syndrome.
\newblock {\em American Journal of Critical Care} {\bf 24,} 15--23.

\bibitem[\protect\citeauthoryear{Shickel, Tighe, Bihorac, and Rashidi}{Shickel
  et~al.}{2018}]{Shickel2018}
Shickel, B., Tighe, P.~J., Bihorac, A., and Rashidi, P. (2018).
\newblock Deep {EHR}: A survey of recent advances in deep learning techniques
  for electronic health record ({EHR}) analysis.
\newblock {\em {IEEE} Journal of Biomedical and Health Informatics} {\bf 22,}
  1589--1604.

\bibitem[\protect\citeauthoryear{Smith and Gelfand}{Smith and
  Gelfand}{1992}]{Smith1992}
Smith, A. F.~M. and Gelfand, A.~E. (1992).
\newblock Bayesian statistics without tears: A sampling{\textendash}resampling
  perspective.
\newblock {\em The American Statistician} {\bf 46,} 84--88.

\bibitem[\protect\citeauthoryear{Stekhoven and Buhlmann}{Stekhoven and
  Buhlmann}{2011}]{Stekhoven2011}
Stekhoven, D.~J. and Buhlmann, P. (2011).
\newblock {MissForest}--non-parametric missing value imputation for mixed-type
  data.
\newblock {\em Bioinformatics} {\bf 28,} 112--118.

\bibitem[\protect\citeauthoryear{Strau{\ss}, Ewig, Richter, König, Heller, and
  Bauer}{Strau{\ss} et~al.}{2014}]{Straus2014}
Strau{\ss}, R., Ewig, S., Richter, K., König, T., Heller, G., and Bauer, T.~T.
  (2014).
\newblock The prognostic significance of respiratory rate in patients with
  pneumonia.
\newblock {\em Deutsches Ärzteblatt international} .

\bibitem[\protect\citeauthoryear{Strauss and Oliva}{Strauss and
  Oliva}{2021}]{Strauss2021}
Strauss, R. and Oliva, J.~B. (2021).
\newblock Arbitrary conditional distributions with energy.
\newblock {\em Advances in Neural Information Processing Systems} {\bf 34,}.

\bibitem[\protect\citeauthoryear{Stubbendick and Ibrahim}{Stubbendick and
  Ibrahim}{2003}]{Stubbendick2003}
Stubbendick, A.~L. and Ibrahim, J.~G. (2003).
\newblock Maximum likelihood methods for nonignorable missing responses and
  covariates in random effects models.
\newblock {\em Biometrics} {\bf 59,} 1140--1150.

\bibitem[\protect\citeauthoryear{Tongyoo, Viarasilpa, and Permpikul}{Tongyoo
  et~al.}{2018}]{Tongyoo2018}
Tongyoo, S., Viarasilpa, T., and Permpikul, C. (2018).
\newblock Serum potassium levels and outcomes in critically ill patients in the
  medical intensive care unit.
\newblock {\em Journal of International Medical Research} {\bf 46,} 1254--1262.

\bibitem[\protect\citeauthoryear{{Tschannen}, {Bachem}, and
  {Lucic}}{{Tschannen} et~al.}{2018}]{Tschannen2018}
{Tschannen}, M., {Bachem}, O., and {Lucic}, M. (2018).
\newblock {Recent Advances in Autoencoder-Based Representation Learning}.
\newblock {\em arXiv e-prints} page arXiv:1812.05069.

\bibitem[\protect\citeauthoryear{Van~Buuren and Groothuis-Oudshoorn}{Van~Buuren
  and Groothuis-Oudshoorn}{2011}]{VanBuuren2011}
Van~Buuren, S. and Groothuis-Oudshoorn, K. (2011).
\newblock mice: Multivariate imputation by chained equations in r.
\newblock {\em Journal of statistical software} {\bf 45,} 1--67.

\bibitem[\protect\citeauthoryear{{Vinzamuri} and {Reddy}}{{Vinzamuri} and
  {Reddy}}{2013}]{Vinzamuri2013}
{Vinzamuri}, B. and {Reddy}, C.~K. (2013).
\newblock Cox regression with correlation based regularization for electronic
  health records.
\newblock In {\em 2013 IEEE 13th International Conference on Data Mining},
  pages 757--766.

\bibitem[\protect\citeauthoryear{Wang, Yao, and Zhao}{Wang
  et~al.}{2016}]{Wang2016}
Wang, Y., Yao, H., and Zhao, S. (2016).
\newblock Auto-encoder based dimensionality reduction.
\newblock {\em Neurocomputing} {\bf 184,} 232--242.

\bibitem[\protect\citeauthoryear{Wells, Nowacki, Chagin, and Kattan}{Wells
  et~al.}{2013}]{Wells2013}
Wells, B.~J., Nowacki, A.~S., Chagin, K., and Kattan, M.~W. (2013).
\newblock Strategies for handling missing data in electronic health record
  derived data.
\newblock {\em {eGEMs} (Generating Evidence {\&} Methods to improve patient
  outcomes)} {\bf 1,} 7.

\end{thebibliography}


\begin{thebibliography}{xx}

\harvarditem{Diggle \harvardand\ Kenward}{1994}{Diggle1994}
Diggle, P. \harvardand\ Kenward, M.~G.  \harvardyearleft 1994\harvardyearright
  , `Informative drop-out in longitudinal data analysis', {\em Applied
  Statistics} {\bf 43}(1),~49.

\harvarditem{Dua \harvardand\ Graff}{2017}{Dua2019}
Dua, D. \harvardand\ Graff, C.  \harvardyearleft 2017\harvardyearright , `{UCI}
  machine learning repository'.
\newline\harvardurl{http://archive.ics.uci.edu/ml}

\harvarditem{Ferreira}{2001}{Ferreira2001}
Ferreira, F.~L.  \harvardyearleft 2001\harvardyearright , `Serial evaluation of
  the {SOFA} score to predict outcome in critically ill patients', {\em {JAMA}}
  {\bf 286}(14),~1754.

\harvarditem[Gall et~al.]{Gall, Loirat, Alperovitch, Glaser, Granthil, Mathieu,
  Mercier, Thomas \harvardand\ Villers}{1984}{JeanRoger1984}
Gall, J.-R.~L., Loirat, P., Alperovitch, A., Glaser, P., Granthil, C., Mathieu,
  D., Mercier, P., Thomas, R. \harvardand\ Villers, D.  \harvardyearleft
  1984\harvardyearright , `A simplified acute physiology score for {ICU}
  patients', {\em Critical Care Medicine} {\bf 12}(11),~975--977.

\harvarditem[Ivanov et~al.]{Ivanov, Figurnov \harvardand\
  Vetrov}{2019}{Ivanov2019}
Ivanov, O., Figurnov, M. \harvardand\ Vetrov, D.  \harvardyearleft
  2019\harvardyearright , Variational autoencoder with arbitrary conditioning,
  {\em in} `International Conference on Learning Representations'.
\newline\harvardurl{https://openreview.net/forum?id=SyxtJh0qYm}

\harvarditem[{Johnson} et~al.]{{Johnson}, {Dunkley}, {Mayaud}, {Tsanas},
  {Kramer} \harvardand\ {Clifford}}{2012}{Johnson2012}
{Johnson}, A. E.~W., {Dunkley}, N., {Mayaud}, L., {Tsanas}, A., {Kramer}, A.~A.
  \harvardand\ {Clifford}, G.~D.  \harvardyearleft 2012\harvardyearright ,
  Patient specific predictions in the intensive care unit using a bayesian
  ensemble, {\em in} `2012 Computing in Cardiology', pp.~249--252.

\harvarditem{{Kingma} \harvardand\ {Welling}}{2013}{Kingma2013}
{Kingma}, D.~P. \harvardand\ {Welling}, M.  \harvardyearleft
  2013\harvardyearright , `{Auto-Encoding Variational Bayes}', {\em arXiv
  e-prints} p.~arXiv:1312.6114.

\harvarditem{Mattei \harvardand\ Frellsen}{2019}{Mattei2019}
Mattei, P.-A. \harvardand\ Frellsen, J.  \harvardyearleft 2019\harvardyearright
  , {MIWAE}: Deep generative modelling and imputation of incomplete data sets,
  {\em in} K.~Chaudhuri \harvardand\ R.~Salakhutdinov, eds, `Proceedings of the
  36th International Conference on Machine Learning', Vol.~97 of {\em
  Proceedings of Machine Learning Research}, PMLR, Long Beach, California, USA,
  pp.~4413--4423.
\newline\harvardurl{http://proceedings.mlr.press/v97/mattei19a.html}

\harvarditem[{Nazabal} et~al.]{{Nazabal}, {Olmos}, {Ghahramani} \harvardand\
  {Valera}}{2018}{Nazabal2018}
{Nazabal}, A., {Olmos}, P.~M., {Ghahramani}, Z. \harvardand\ {Valera}, I.
  \harvardyearleft 2018\harvardyearright , `{Handling Incomplete Heterogeneous
  Data using VAEs}', {\em arXiv e-prints} p.~arXiv:1807.03653.

\harvarditem{Prechelt}{1998}{Prechelt1998}
Prechelt, L.  \harvardyearleft 1998\harvardyearright , Early stopping-but
  when?, {\em in} `Neural Networks: Tricks of the trade', Springer, pp.~55--69.

\harvarditem[Saxe et~al.]{Saxe, Mcclelland \harvardand\
  Ganguli}{2014}{Saxe2014}
Saxe, A.~M., Mcclelland, J.~L. \harvardand\ Ganguli, S.  \harvardyearleft
  2014\harvardyearright , Exact solutions to the nonlinear dynamics of learning
  in deep linear neural network, {\em in} `In International Conference on
  Learning Representations'.

\harvarditem[Silva et~al.]{Silva, Moody, Scott, Celi \harvardand\
  Mark}{2012}{Silva2012}
Silva, I., Moody, G., Scott, D.~J., Celi, L.~A. \harvardand\ Mark, R.~G.
  \harvardyearleft 2012\harvardyearright , `Predicting in-hospital mortality of
  icu patients: The physionet/computing in cardiology challenge 2012.', {\em
  Computing in cardiology} {\bf 39},~245--248.

\end{thebibliography}

\section*{Supporting Information}
"Web Appendices A and B, referenced in Sections \ref{sec:meth} - \ref{sec:examples2} are available with this paper at the Biometrics website on Wiley Online Library." Additionally, the NIMIWAE R package containing code to perform the diagnostic methods described in the article can be downloaded from \url{https://github.com/DavidKLim/NIMIWAE}, and the repository of code to replicate all figures and tables from this paper can be found here: \url{https://github.com/DavidKLim/NIMIWAE_Paper}.

%


\label{lastpage}
\end{document}


\maketitle

\setstretch{2}


\section{Web Appendix A: Additional Details}

\subsection{IWAE Architecture}
\begin{figure}[H]
\begin{center}
\includegraphics[width=130mm]{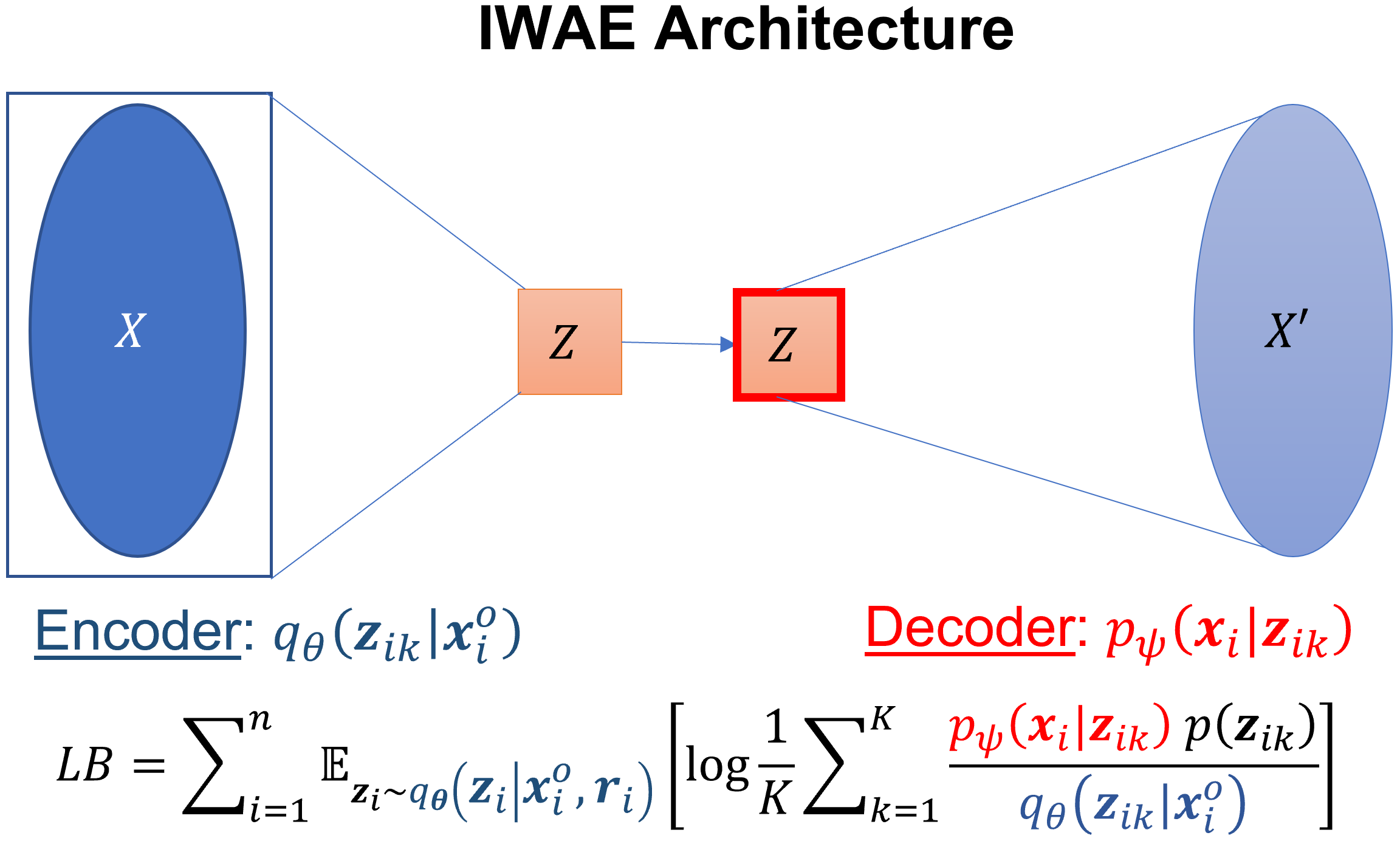} 
\end{center}
\caption{\small Architecture of an importance weighted autoencoder (IWAE) in the absence of missing data. Darkly colored nodes represent deterministic values, lightly colored nodes represent learned distributional parameters, and outlined (in red) nodes represent sampled values from learned distributions. Orange cells correspond to latent variables $\mathbf{Z}$. $\mathbf{Z}_1,\ldots,\mathbf{Z}_K$ is sampled $1$ time each from the variational posterior posterior distribution $q(\mathbf{Z}|\mathbf{X})$. Below is the lower bound ($LB$), which is optimized via stochastic gradient descent.}
\label{fig:IWAEarchitecture}
\end{figure}

\subsection{VAEs and IWAEs with Ignorable Missingness} \label{sec:VAEIgnorable}
There are a number of VAE/IWAE methods that have been developed to handle ignorably missing data. 



\pkg{VAEAC} \citep{Ivanov2019} is a method catered for datasets with complete training sets. \pkg{VAEAC} can handle different types of data (continuous, categorical, or count), and additionally conditions on the missingness mask. In training time, additional MCAR missingness is imposed on the fully-observed training set by masking a fixed proportion of values. This trains the neural network to learn to impute missing values accurately during testing. Missing data is imputed by values that are sampled from the generative distribution $p(\mathbf{X}^m|\mathbf{Z},\mathbf{X}^o,\mathbf{R})$. However, inducing additional missingness may be problematic for situations where there already exists a high level of inherent missingness in the data.

\pkg{HI-VAE} \citep{Nazabal2018} takes advantage of the fact that the marginal log-likelihood simplifies to Equation (3) of the main text under ignorable missingness. Under this simplification, the new ELBO corresponding to Equation (1) of the main text is given by
\begin{equation}
\mathcal{L}(\theta,\psi)= \E_{\mathbf{Z} \sim q_\theta(\mathbf{Z}|\mathbf{X}^o)} \log \left[ \frac{p_\psi(\mathbf{X}^o|\mathbf{Z})p(\mathbf{Z})}{q_\theta(\mathbf{Z}|\mathbf{X}^o)} \right],
\label{eqn:ELBOIgnorable}
\end{equation}
such that the training of the VAE depends only on the observed values. To preserve the length of each observation vector, the missing features are pre-imputed by some value, typically using zero imputation. In this setting, the standard variational approximation $q_{\theta}(\mathbf{Z}|\mathbf{X})$ is replaced by $q_{\theta}(\mathbf{Z}|\mathbf{X}^o)$. This quantity approximates $p(\mathbf{Z}|\mathbf{X}^o)$ instead of $p(\mathbf{Z}|\mathbf{X})$, thus allowing the VAE to be trained with partially-observed input data. In \pkg{HI-VAE}, the VAE framework is adapted to be able to deal with heterogenous data, i.e. both discrete and continuous data. Missing data is imputed by either sampling from the generative model $p(\mathbf{X}^m|\mathbf{Z})$, or by setting missing values equal to the corresponding means of the posterior distribution. In contrast, the \pkg{MIWAE} method \citep{Mattei2019} does not take heterogeneous data as input, but uses the IWAE framework to create a tighter lower bound on the marginal log-likelihood. They also use a principled imputation scheme that works up to MAR missingness, utilizing importance-weighted samples to calculate $\E[\mathbf{X}^m | \mathbf{X}^o,\mathbf{Z}]$.

\subsection{NIMIWAE Training Algorithm}

The training of the NIMIWAE architecture proceeds as follows:
\begin{enumerate}
\item The missing entries are pre-imputed to zero and appended with observed entries, and fed into the encoder, or $g_{\theta_1}(\mathbf{x}_i^o)$, to learn parameters of $q_{\theta_1}(\mathbf{z}_i|\mathbf{x}_i^o)$. One can also specify a mean pre-imputation, but the choice of the pre-imputation method has been shown to not significantly affect performance in an IWAE network \citep{Mattei2019}.
\item $K$ samples are drawn from $q_{\theta_1}(\mathbf{z}_i|\mathbf{x}_i^o)$.
\item Samples from (2) are used as input for the decoder, or $f_\psi(\mathbf{z}_i)$, to learn the parameters of $p_{\psi}(\mathbf{x}_i|\mathbf{z}_i)$.
\item The samples from (2) are used again as input for the missing data posterior network, or $g_{\theta_2}(\mathbf{z}_i,\mathbf{x}_i^o,\mathbf{r}_i)$, concatenated with the observed data entries (with missing entries pre-imputed to 0) and the missingness mask, to learn parameters of $q_{\theta_2}(\mathbf{x}_i^m|\mathbf{z}_i,\mathbf{x}_i^o,\mathbf{r}_i)$.
\item We draw samples of $\mathbf{x}_i^m$ from $q_{\theta_2}(\mathbf{x}_i^m|\mathbf{z}_i,\mathbf{x}_i^o,\mathbf{r}_i)$, and use them as input, concatenated with the fixed observed entries $\mathbf{x}_i^o$, into missingness network, or $h_\phi(\mathbf{x}_i)$ to learn the parameters associated with the model of the missingness mask $p_\phi(\mathbf{r}_i|\mathbf{x}_i)$.
\end{enumerate}

In NIMIWAE, the neural network that models the missingness, or $h_\phi(\boldsymbol{x}_i)$, contains $p_{miss}$ output nodes, and applies the Sigmoid activation function to the output node to yield probabilities of each partially-observed feature $j_m$ being observed for the $i^{th}$ sample. By default, this network takes all $p$ features as input, and the number of hidden layers and nodes per hidden layer are tuned separately from the rest of the architecture. Each output node in $h_\phi(\boldsymbol{x}_i)$ is exactly equivalent to the logistic regression model proposed by \citet{Diggle1994} (as given in Section 2.3.2 of the main text) when $h_\phi(\mathbf{x}_i)$ contains no hidden layers, and the selection of additional hidden layers can allow for the capturing of more complex effects. 

Under simple distributional assumptions of $q_{\theta_2}(\mathbf{x}_i^m|\mathbf{z}_i,\mathbf{r}_i,\mathbf{x}_i^o)$, the sampling step in Step (4) is similar to the sampling of the latent variable $\mathbf{z}_i$ in Step (2), and both can be accomplished using the reparametrization trick \citep{Kingma2013}.

\subsubsection{NIMIWAE: Initialization, Early Stop, and Hyperparameter Tuning}

Initialization of weights and biases in deep learning architectures can significantly affect the trained model, especially in datasets with smaller sample sizes. By default, NIMIWAE uses the semi-orthogonal matrix initialization \citep{Saxe2014} for $\psi$, $\theta$, and $\phi$. Alternatively, we initialize weights pertaining to the missing features in the input layer of the missingness network using values drawn from a Uniform(-2,2) distribution. This is done in order to draw larger initial values of the effects of missing variables on the missingness, in order to encourage the network to learn nonzero effects of these missing features. Empirically, we found that this alternative initialization helps the network impute more accurate values in smaller sample size settings, particularly under MNAR missingness, while maintaining similar performance in the MCAR and MAR cases. 

We also incorporate an early stop criterion \citep{Prechelt1998} in order to prevent overfitting on the training set, and to reduce computation time. Specifically, let $L^{(\tau)} \equiv \hat{\mathcal{L}}_{K,valid}^{NIMIWAE, (\tau)}$ denote the estimated NIMIWAE bound on a held-out validation set at each epoch $(\tau)$, and initialize $L_{opt} = L^{(0)}$ and $E^{(0)} = 0$. Each epoch consists of $\lceil n_{train}/bs \rceil$ updates, where $n_{train}$ is the number of observations in the training set and $bs$ is the mini-batch size hyperparameter, and $\lceil x \rceil$ is the smallest integer greater than or equal to $x$. At each epoch, the $n_{train}$ observations are divided into approximately equal size mini-batches of at most $bs$ observations each, and updates of the neural network parameters are done on by estimating the NIMIWAE bound on each mini-batch, cycling through each mini-batch such that all observations are involved in the updates for during each epoch. For $\tau\geq 1$, if $L^{(\tau)} - L_{opt} > 0$, we replace $L_{opt} = L^{(\tau)}$. Also, if $L^{(\tau)} - L_{opt} \leq \varepsilon L_{opt}$, we set $E^{(\tau)}=E^{(\tau-1)}+1$. In this way, training continues as long as the improvement in the estimated lower bound in the validation set is greater than $\varepsilon L_{opt}$, but if not, training is allowed to run for a set number of epochs before early stopping. This leeway (called ``$patience$'') is allowed due to the properties of SGD, which may cause $L^{(\tau)}$ to fluctuate, especially for smaller $\tau$. If $E^{(\tau)} = patience$, we save the optimal model pertaining to $L_{opt}$, and terminate training. Here, we set $patience=50$ and $\varepsilon=0.0001$.

In NIMIWAE, we have several hyperparameters that need to be tuned (default values in parentheses): 1) learning rate $(\{0.001,0.01\})$, 2) number of hidden layers $(\{0,1,2\})$ and nodes per hidden layer $(\{64,128\})$, 3) dimensionality of $\mathbf{Z}$ $(\{\lfloor p/12 \rfloor , \lfloor p/4 \rfloor , \lfloor p/2 \rfloor , \lfloor 3p/4 \rfloor \})$, where $\lfloor x \rfloor$ is the largest integer less than or equal to $x$. We also tune (2) separately for the missingness network $(\{0,1\}$ and $\{16, 32\})$ ($h_\phi(\cdot)$) vs. the rest of the networks ($f_\psi(\cdot),g_{\theta_1}(\cdot),g_{\theta_2}(\cdot)$). In order to find the optimal combination of values for these hyperparameters, we perform an extensive grid search of all combinations of prespecified values for these quantities. One may also choose to tune the number of latent variables $K$. In this paper, we fix $K=5$ during training for computational ease.

\section{Web Appendix B: Additional Simulations and Details}
\subsection{Additional Simulations}
\begin{figure}
\begin{center}
\includegraphics[width=175mm]{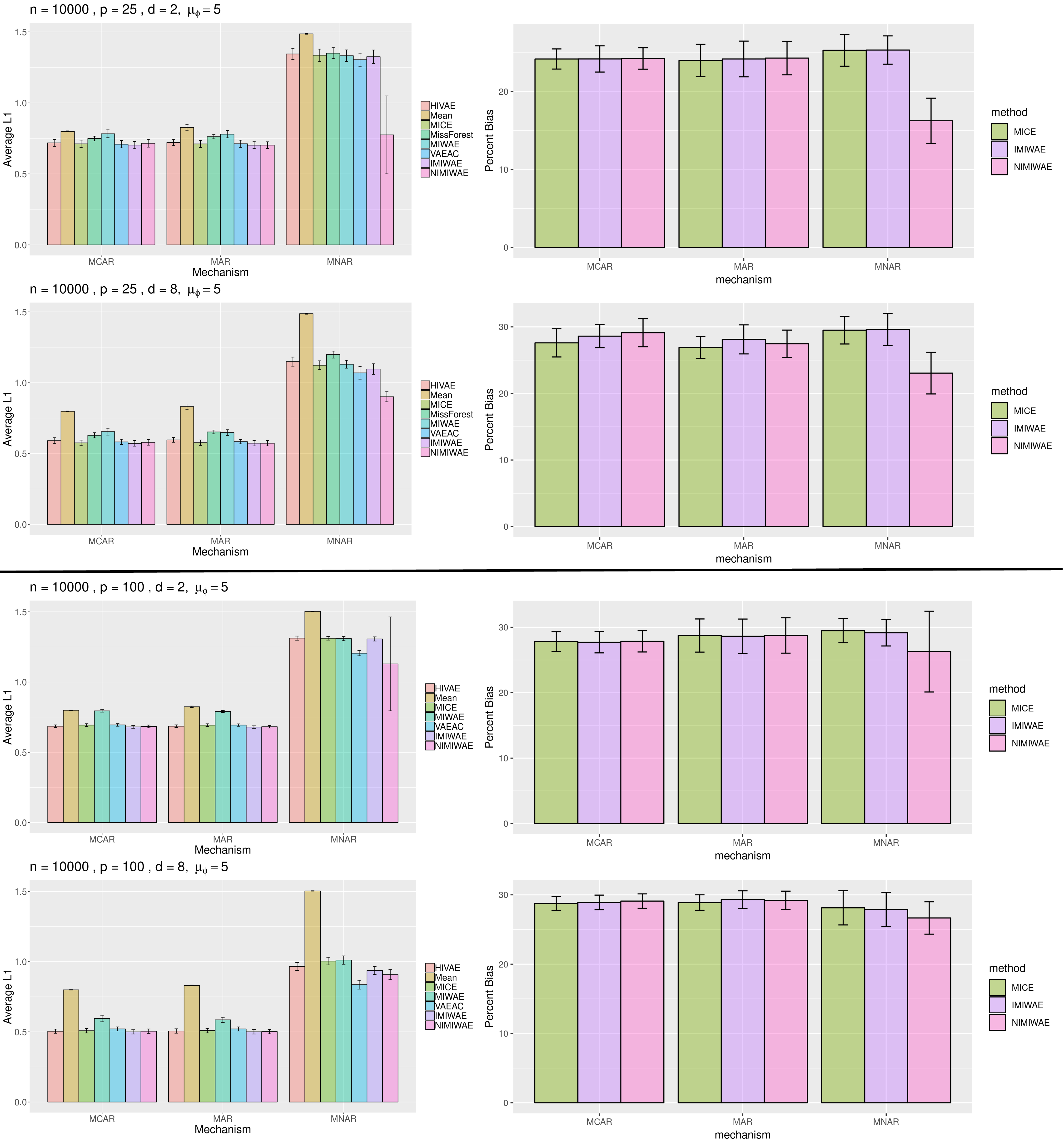} 
\end{center}
\caption{\small Average L1 distance between true and imputed values for missing entries stratified by $d$ (left), and percent bias of pooled coefficient estimates. NIMIWAE outperforms all methods in imputing missing values that were simulated to be MNAR  when $p=25$, but performs poorly under $p=100$. Here, $n=10,000$, $\mu_\phi=5$, and error bars show the variability of each metric across the 5 reps. NIMIWAE struggles to handle the difficult MNAR missingness pattern when the sample size is smaller, while the dimensionality of the data is large.}
\label{fig:simsN10K}
\end{figure}

\begin{figure}
\begin{center}
\includegraphics[width=175mm]{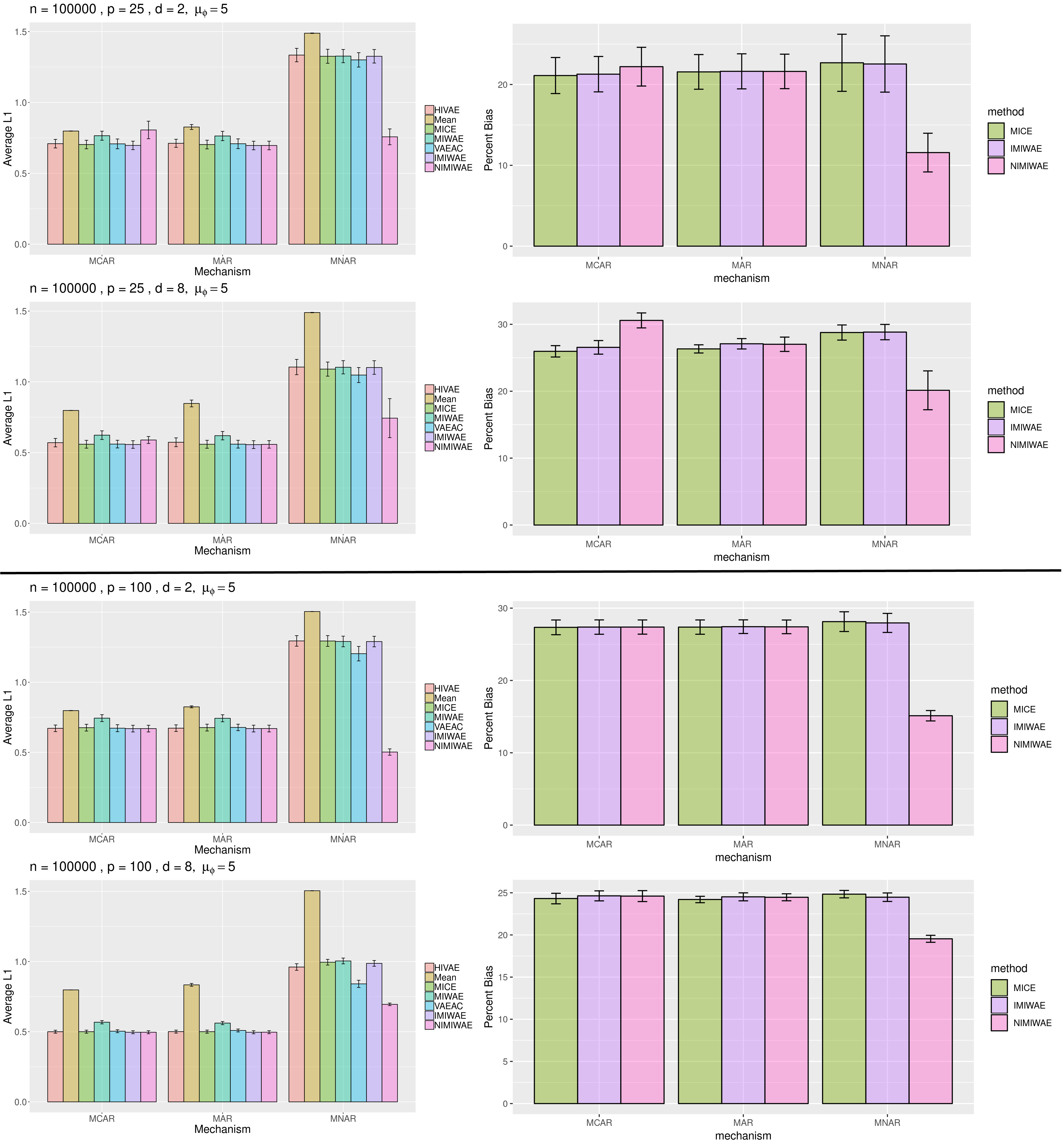} 
\end{center}
\caption{\small Average L1 distance between true and imputed values for missing entries stratified by $d$ (left), and percent bias of pooled coefficient estimates. NIMIWAE was run using the alternative initialization method, as described in Section 2.3 of the main text. Again, NIMIWAE outperforms all methods in imputing missing values that were simulated to be MNAR. Here, $n=100,000$, $\mu_\phi=5$, and error bars show the variability of each metric across the 5 reps.}
\label{fig:simsN100K_alt}
\end{figure}


Web Figure \ref{fig:simsN10K} shows results from the simulation setup with $n=10,000$, and Web Figure \ref{fig:simsN100K_alt} shows results from the alternate setup with $n=100,000$, varying $p=\{25,100\}$ and $d=\{2,8\}$ in each case.

As in all deep learning algorithms, NIMIWAE performs best under larger sample sizes, and this aspect is especially important given the additional task in NIMIWAE of narrowing down relevant features in a potentially overspecified missingness model. When a large number of samples is not readily available, and the dimensionality of the data is very large, one can either utilize an alternative method of weight initialization for the missing features in the missingness network, as seen in Section 3.1 of the main text, or narrow down the number of features that are included in NIMIWAE's missingness network using some prior knowledge on the dependencies of the features with the missingness. We found that under the large sample size setting ($n=100,000$), imputation performance was generally robust to the method of initialization, as shown in Web Figure \ref{fig:simsN100K_alt}. We additionally found that having fewer unnecessary features in the missingness model greatly increased the performance of both imputation and inference under MNAR, as one would expect. However, omitting a feature that is truly relevant to the missingness may cause very poor performance in both imputation and downstream coefficient estimation, and in practice, determining which features are relevant to the missingness may be nontrivial. One may utilize \textit{a priori} knowledge about the data in order to inform a decision on which features are significant in the missingness model, but such decisions may not be able to be made with a high levels of certainty.


\subsection{Tuning Hyperparameters}
In this section, we summarize the combinations of hyperparameter values that were searched. For each method and dataset, we searched over combinations of two different values for each hyperparameter that was tuned. We tuned several hyperparameters for each method: number of nodes per hidden layer ($h$), number of nodes per hidden layer in the missingness network ($h_r$) , number of hidden layers ($nhl$), number of hidden layers in the missingness network ($nhl_r$), dimensionality of latent space ($dz$), and learning rate ($lr$). For HI-VAE, $nhl$ was not able to be varied. Therefore, we fixed $nhl$ and tuned the dimensionality of the $y$ latent variable ($dy$) instead. Hyperparameters $nhl_r$ and $h_r$ were applicable only to NIMIWAE, so they were tuned only for NIMIWAE. Hyperparameters batch size ($bs$) and maximum number of epochs ($epochs_{max}$) were fixed in our analyses. Below are the searched values of the hyperparameters for each dataset in our analyses:

\begin{itemize}
  \item $p=(25, 100)$ Simulated data with $n=(10,000; 100,000)$
  \begin{itemize}
    \item $h=\{128,64\}$, $h_r=\{16,32\}$
    \item $lr=\{0.001,0.01\}$
    \item $dz=\{\lfloor 3*p/4 \rfloor, \lfloor p/2 \rfloor, \lfloor p/4 \rfloor, 8\}$
    \item $nhl=\{0,1,2\}$
    \item $nhl_r=\{0,1\}$
    \item $bs=1,000$ for $n=10,000$ and $bs=10,000$ for $n=100,000$, $epochs_{max}=2002$
  \end{itemize}
  \item banknote
  \begin{itemize}
    \item $h=\{128,64\}$, $h_r=\{16,32\}$
    \item $lr=\{0.001,0.01\}$
    \item $dz=\{1,2,3\}$
    \item $nhl=\{0,1,2\}$
    \item $nhl_r=\{0,1\}$
    \item $bs=200$, $epochs=2002$
  \end{itemize}
  \item concrete
  \begin{itemize}
    \item $h=\{128,64\}$, $h_r=\{16,32\}$
    \item $lr=\{0.001,0.01\}$
    \item $dz=\{2,4,6\}$
    \item $nhl=\{0,1,2\}$
    \item $nhl_r=\{0,1\}$
    \item $bs=200$, $epochs=2002$
  \end{itemize}
  \item hepmass
  \begin{itemize}
    \item $h=\{128,64\}$, $h_r=\{16,32\}$
    \item $lr=\{0.001,0.01\}$
    \item $dz=\{5,10,15\}$
    \item $nhl=\{0,1,2\}$
    \item $nhl_r=\{0,1\}$
    \item $bs=10,000$, $epochs_{max}=2002$
  \end{itemize}
  \item power
  \begin{itemize}
    \item $h=\{128,64\}$, $h_r=\{16,32\}$
    \item $lr=\{0.001,0.01\}$
    \item $dz=\{1,3,4\}$
    \item $nhl=\{0,1,2\}$
    \item $nhl_r=\{0,1\}$
    \item $bs=10,000$, $epochs_{max}=2002$
  \end{itemize}
  \item red
  \begin{itemize}
    \item $h=\{128,64\}$, $h_r=\{16,32\}$
    \item $lr=\{0.001,0.01\}$
    \item $dz=\{2,5,8\}$
    \item $nhl=\{0,1,2\}$
    \item $nhl_r=\{0,1\}$
    \item $bs=200$, $epochs_{max}=2002$
  \end{itemize}
  \item white
  \begin{itemize}
    \item $h=\{128,64\}$, $h_r=\{16,32\}$
    \item $lr=\{0.001,0.01\}$
    \item $dz=\{2,5,8\}$
    \item $nhl=\{0,1,2\}$
    \item $nhl_r=\{0,1\}$
    \item $bs=200$, $epochs_{max}=2002$
  \end{itemize}
  \item Physionet 2012 Challenge data
  \begin{itemize}
    \item $h=\{128,64\}$, $h_r=\{16,32\}$
    \item $lr=\{0.001,0.01\}$
    \item $dz=\{8, 28, 57, 85\}$
    \item $nhl=\{0,1,2\}$
    \item $nhl_r=\{0,1,2\}$
    \item $bs=1,000$, $epochs_{max}=2002$
  \end{itemize}
\end{itemize}

\section{Web Appendix C: Data}
Datasets used for analysis in this study are all publicly available online. Details of how to obtain these datasets are given in this section.
\subsection{UCI Machine Learning Datasets}
The UCI datasets can be obtained from the UCI Machine Learning repository at \url{https://archive.ics.uci.edu/ml/index.php} \citep{Dua2019} The datasets included in the analyses in this study are given below:
\begin{enumerate}
  \item banknote: $n=1372$ and $p=4$.
    \begin{itemize}
      \item Documentation: \url{https://archive.ics.uci.edu/ml/datasets/banknote+authentication}
      \item Link to Data: \url{https://archive.ics.uci.edu/ml/machine-learning-databases/00267/data_banknote_authentication.txt}
      \item Pre-processing: Removed class variable (last feature)
    \end{itemize}
  \item concrete: $n=1030$ and $p=8$
    \begin{itemize}
      \item Documentation: \url{https://archive.ics.uci.edu/ml/datasets/concrete+compressive+strength}
      \item Link to Data: \url{https://archive.ics.uci.edu/ml/machine-learning-databases/concrete/compressive/Concrete_Data.xls}
      \item Pre-processing: Removed output variable ``Concrete compressive strength" (last feature)
    \end{itemize}
  \item hepmass: $n=300,000$ and $p=21$
    \begin{itemize}
      \item Documentation: \url{https://archive.ics.uci.edu/ml/datasets/HEPMASS}
      \item Link to Data: \url{https://archive.ics.uci.edu/ml/machine-learning-databases/00347}
      \item Pre-processing: We followed the same steps here: \url{https://github.com/gpapamak/maf/blob/master/datasets/hepmass.py}
    \end{itemize}
  \item power: $n=1,000,000$ and $p=6$
    \begin{itemize}
      \item Documentation: \url{https://archive.ics.uci.edu/ml/datasets/Individual+household+electric+power+consumption}
      \item Link to Data: \url{https://archive.ics.uci.edu/ml/machine-learning-databases/00235/household_power_consumption.zip}
      \item Pre-processing: We followed the same steps here: \url{https://github.com/gpapamak/maf/blob/master/datasets/power.py}
    \end{itemize}
  \item red: $n=1599$ and $p=11$
    \begin{itemize}
      \item Documentation: \url{https://archive.ics.uci.edu/ml/datasets/wine+quality}
      \item Link to Data: \url{https://archive.ics.uci.edu/ml/machine-learning-databases/wine-quality/winequality-red.csv}
      \item Pre-processing: Removed wine quality score (last feature)
    \end{itemize}
  \item white: $n=4898$ and $p=11$
    \begin{itemize}
      \item Documentation: \url{https://archive.ics.uci.edu/ml/datasets/wine+quality}
      \item Link to Data: \url{https://archive.ics.uci.edu/ml/machine-learning-databases/wine-quality/winequality-white.csv}
      \item Pre-processing: Removed wine quality score (last feature)
    \end{itemize}
\end{enumerate}

\subsection{Physionet 2012 Challenge Dataset}

The Physionet 2012 Challenge dataset contains ICU measurements from 12,000 de-identified patients admitted at Beth Israel Deconess Medical Center from 2001 and 2008. The dataset is publicly available here: \url{https://physionet.org/content/challenge-2012/1.0.0/}. Patients whose ICU stays were less than 48 hours long were excluded from the dataset. For each patient, 36 features were measured, with one potential measurement taken hourly for 48 hours. Example features include albumin levels (in g/dL), serum sodium (in mEq/L) and white blood cell count (in cells/nL).  For each patient, clinical outcomes such as the SAPS-I score \citep{JeanRoger1984}, SOFA score \citep{Ferreira2001}, length of stay, survival, and in-hospital death were also recorded. The challenge called for learning algorithms to accurately predict the outcome of interest (like in-hospital death) based on such clinical data, facilitating clinical decision-making regarding the need for early intervention in ICU patients \citep{Silva2012}.

The original challenge dataset was divided into training, validation, and test datasets of equal sizes, and missingness was highly prevalent in all three datasets. Table \ref{tab:descriptive} shows the missingness percentage of all features in the dataset, and the number of subjects with at least one non-missing entry recorded for each respective feature.

\begin{table}[ht]
\centering
\begingroup\fontsize{11pt}{9pt}\selectfont
\begin{tabular}{lp{1.5in}cc}
  \hline
 & {\footnotesize Feature} & {\footnotesize Missingness (overall)} & {\footnotesize Patients with $\geq 1$ obs.} \\
  \hline
1 & ALP & 0.984 & 0.425 \\
\rowcolor[gray]{0.90}2 & ALT & 0.983 & 0.434 \\
   3 & AST & 0.983 & 0.434 \\
    \rowcolor[gray]{0.90}4 & Albumin & 0.987 & 0.406 \\
   5 & BUN & 0.928 & 0.986 \\
    \rowcolor[gray]{0.90}6 & Bilirubin & 0.983 & 0.435 \\
   7 & Cholesterol & 0.998 & 0.079 \\
    \rowcolor[gray]{0.90}8 & Creatinine & 0.927 & 0.986 \\
   9 & DiasABP & 0.458 & 0.703 \\
    \rowcolor[gray]{0.90}10 & FiO2 & 0.843 & 0.677 \\
   11 & GCS & 0.680 & 0.986 \\
    \rowcolor[gray]{0.90}12 & Glucose & 0.932 & 0.976 \\
   13 & HCO3 & 0.929 & 0.984 \\
    \rowcolor[gray]{0.90}14 & HCT & 0.905 & 0.985 \\
   15 & HR & 0.098 & 0.986 \\
    \rowcolor[gray]{0.90}16 & K & 0.924 & 0.980 \\
   17 & Lactate & 0.959 & 0.548 \\
    \rowcolor[gray]{0.90}18 & MAP & 0.461 & 0.701 \\
   19 & Mg & 0.929 & 0.977 \\
    \rowcolor[gray]{0.90}20 & MechVent & 0.849 & 0.632 \\
   21 & NIDiasABP & 0.579 & 0.875 \\
    \rowcolor[gray]{0.90}22 & NIMAP & 0.585 & 0.873 \\
   23 & NISysABP & 0.579 & 0.877 \\
    \rowcolor[gray]{0.90}24 & Na & 0.929 & 0.983 \\
   25 & PaCO2 & 0.885 & 0.755 \\
    \rowcolor[gray]{0.90}26 & PaO2 & 0.885 & 0.755 \\
   27 & Platelets & 0.926 & 0.984 \\
    \rowcolor[gray]{0.90}28 & RespRate & 0.759 & 0.278 \\
   29 & SaO2 & 0.960 & 0.448 \\
    \rowcolor[gray]{0.90}30 & SysABP & 0.458 & 0.703 \\
   31 & Temp & 0.629 & 0.986 \\
    \rowcolor[gray]{0.90}32 & TroponinI & 0.998 & 0.047 \\
   33 & TroponinT & 0.989 & 0.220 \\
    \rowcolor[gray]{0.90}34 & Urine & 0.307 & 0.975 \\
   35 & WBC & 0.933 & 0.983 \\
    \rowcolor[gray]{0.90}36 & pH & 0.879 & 0.760 \\
    \hline
 \end{tabular}
 \endgroup
 \caption{Proportion of overall missingness for each feature, and proportion of patients with at least 1 non-missing measurement for each feature in the Physionet 2012 Challenge EHR dataset.}
 \label{tab:descriptive}
 \end{table}
 
The data was pre-processed following \citet{Johnson2012}, and this procedure, as well as a copy of this data can be found here: \url{https://github.com/alistairewj/challenge2012}. Domain knowledge and distributional assumptions were used in order to convert implausible values, and temporal data was processed by using domain knowledge to break down each feature by summary statistics: first, last, lowest, highest, and median values across the time points. The resulting dataset contained 114 features for each of the 12,000 patients. In this way, a measurement is missing for a patient only if the diagnostic measure was not taken during any of the 48 time points. We also split the patients with the same 8:2 training-validation set ratio as in the simulations. For each method, we trained each imputation model on the training set, and imputed the missing values in the training set. The validation set was used only to select optimal hyperparameter values for deep learning models: NIMIWAE, IMIWAE, HIVAE, VAEAC, and MIWAE.

\bibliographystyle{agsm}
\bibliography{P2}